%% file: smh_paper.tex
\documentclass[11pt]{elsarticle}

\usepackage{lineno,hyperref}

\journal{}

\let\today\relax
\makeatletter
\def\ps@pprintTitle{%
    \let\@oddhead\@empty
    \let\@evenhead\@empty
    \def\@oddfoot{\footnotesize\itshape
         {} \hfill\today}%
    \let\@evenfoot\@oddfoot
    }
\makeatother

\usepackage[utf8]{inputenc}
\usepackage{algorithm}
\usepackage[noend]{algpseudocode}
\usepackage{multirow}
\usepackage{booktabs}
\usepackage{graphicx}
\usepackage{subfig}
\usepackage{svg} 
\usepackage{comment}
\usepackage{verbatim}
\usepackage{color}
\usepackage{makecell}
\usepackage{amsmath,amssymb}

\usepackage{setspace}
\usepackage{geometry}

\usepackage{hyperref}
\hypersetup{
 colorlinks=true,
 linkcolor=darkgreen,
 filecolor=darkgreen,
 citecolor = darkgreen,      
 urlcolor=cyan,
}

\newcommand{\tcolB}{\textcolor{black}}

\newcommand{\ignore}[1]{}

\algnewcommand{\LineCommentt}[1]{\State \(\triangleright\) #1}

\begin{document}

\begin{frontmatter}

\title{Assessing the impact of emergency department short stay units using length-of-stay prediction and discrete event simulation}



\fntext[myfootnote]{Joint senior authors}

\author[dsl_address]{Mucahit Cevik}
\ead{mcevik@torontomu.ca}

\author[dsl_address]{Can Kavaklioglu}
\ead{can.kavaklioglu@torontomu.ca}



\author[st_michaels_address]{Fahad Razak}
\ead{fahad.razak@mail.utoronto.ca}

\author[st_michaels_address]{Amol Verma\fnref{myfootnote}}
\ead{amol.verma@mail.utoronto.ca}

\author[dsl_address]{Ayse Basar\fnref{myfootnote}}
\ead{ayse.bener@torontomu.ca}

\address[dsl_address]{Data Science Lab, Department of MIE, Toronto Metropolitan University, Canada}
\address[st_michaels_address]{Li Ka Shing Knowledge Institute, St. Michaels Hospital, Unity Health Toronto, University of Toronto, Canada}

\begin{abstract}
Accurately predicting hospital length-of-stay at the time a patient is admitted to hospital may help guide clinical decision making and resource allocation. In this study we aim to build a decision support system that predicts hospital length-of-stay for patients admitted to general internal medicine from the emergency department. We conduct an exploratory data analysis and employ feature selection methods to identify the attributes that result in the best predictive performance. We also develop a discrete-event simulation model to assess the performances of the prediction models in a practical setting. Our results show that the recommendation performances of the proposed approaches are generally acceptable and do not benefit from the feature selection. Further, the results indicate that hospital length-of-stay could be predicted with reasonable accuracy (e.g., AUC value for classifying short and long stay patients is 0.69) using patient admission demographics, laboratory test results, diagnostic imaging, vital signs and clinical documentation.  
\end{abstract}


\end{frontmatter}

\nolinenumbers
\section{Introduction}
Hospital length-of-stay (LOS) is an important measure of healthcare resource consumption and is associated with various clinical outcomes. 
Because hospitals commonly operate near or above capacity, safely reducing hospital LOS is an important health system priority. 
Moreover, patients may spend a long time in the emergency department (ED) waiting for an inpatient bed to become available. 
Thus, improving patient flow through the ED and reducing wait times have been a focus for the hospitals \citep{waitTimes2016}.

Typically, patients requiring emergency hospital care are first assessed by a triage nurse upon arrival to the ED. 
They are subsequently examined by an ED physician, who may order a number of tests and treatments. 
The ED physicians may involve a specialist consultant if they feel that the patient will require ongoing hospital care. 
If admitted, the patient waits in the ED until a bed becomes available in one of the hospital wards. 
The average waiting time of patients admitted to hospital through this process is approximately 15 hours \citep{verma2019characteristics}.

One potential strategy for improving patient flow is to stream patients to different hospital units based on predicted LOS. 
Patients with anticipated short hospital LOS (typically less than 48 or 72 hours) could be cared without necessarily admitting them to the hospital, or admitting them to a separate area of the hospital where tests and treatments can be expedited and delivered more efficiently to facilitate rapid discharge.
On the other hand, patients with an anticipated longer duration of hospitalization (e.g., more than 72 hours) could be cared for in conventional hospital wards. 

The problem of classifying patients into short versus longer hospital stays is relatively new, and it is inspired by the proliferation of ``short stay'' or ``observation'' units across North America, Europe, and Australasia \citep{galipeau2015effectiveness, russell2014general, shetty2015age, strom2017hospitalisation, verma2017patient}.
The overall effectiveness of these units remains uncertain~\citep{galipeau2015effectiveness}, however, an analysis of American hospitals found that short stay units could save \$1,572 in costs per patient and that approximately one-third of US hospitals have such a program~\citep{baugh2012making}. 
Short stay units require sophisticated and rapid triaging of patients. 
This is currently performed on an ad hoc basis by nurses or physicians. 
Such ad hoc processes are prone to error and bias, and may result in inefficiency and inequity. 
Studies have shown that, at the time of admission, physicians are able to accurately predict whether a patient will have a short length of stay in only 50\% of the cases~\citep{dent2007can}.
These findings together suggest that classifying patients into short and long predicted LOS is important for hospitals all around the world, is currently performed in a relatively ad hoc way through either clinical judgement or clinical criteria which may not be effective predictors, and highlight an opportunity for data-driven approaches to improve the streamlining of patients for short stay/observation units.

Accurate prediction of patient LOS is important to effectively employ streamlined approaches for assigning the patients to short-stay units or hospital wards. 
In clinical practice, these predictions are often made on an ad hoc basis by clinicians based on their personal experiences without using the information available through prediction models. 
On the other hand, the data that is collected from the patients' medical records, and initial tests/examinations in the ED make it possible to design models for predicting LOS. 
Specifically, after the initial tests and physician examination in the ED, a patient's LOS value can be predicted. 
If the predicted LOS value is smaller than the predetermined LOS threshold (e.g., 72 hours), then the patient can be assigned to short stay units. 
Otherwise, the patient will be designated for a general ward bed, and if there is no available bed, he/she will be temporarily assigned to a bed in the waiting area (or to a bed in short stay units). 
In the case of perfect prediction accuracy, this approach has potential to substantially contribute to quality of care and improved patient flows, since a higher availability of general ward beds will be ensured for long stay patients who typically need a better care. 
In addition, the congestion in the ED will be reduced thanks to improved patient flow. 
We note that reduced prediction accuracy for LOS will inevitably cause a drop in the efficacy of the overall system. Accordingly, we focus on examining the impact of prediction quality for LOS values on hospital operations.

Many factors influence hospital LOS, including severity of patient illness, wait times for tests and procedures, and availability of a discharge destination after hospital (e.g., bed availability in rehabilitation facilities). 
Data pertaining to some of these factors are readily available in hospital electronic medical records. 
The purpose of this study is to use data that are typically available in electronic medical records to develop a predictive model that identifies patients who are expected to have a short hospital LOS. 
We specifically consider general internal medicine (GIM) patients who arrive to the hospital through ED, and use all the data collected through patient's ED stay for the prediction of patient LOS. 
We further propose a discrete-event simulation model to evaluate how a prediction model could be incorporated into the clinical setting. 


The rest of the paper is organized as follows: Section~\ref{sec:litReview} provides a summary of prior work in this area.
Section~\ref{sec:methods} details the proposed methods and Section~\ref{sec:data} describes the data, elaborates on the exploratory analysis, feature selection and experimental design.
The results of the analysis are provided in Section~\ref{sec:results}, which is followed by the discussion and the future directions in Section~\ref{sec:conclusion}.

\section{Literature Review}\label{sec:litReview}
There is an extensive literature pertaining to predicting ED and hospital LOS. We refer the readers to recent review papers on LOS prediction by \citet{awad2017patient} and \citet{lu2015systematic}, and provide a literature review on most relevant studies below.

\citet{rathlev2007time} conducted a time series analysis to examine the factors related to the daily mean LOS in an ED. Their dataset included 93,274 ED visits, and they assessed the independent variables in their model using autoregressive moving average time series analysis. In another study, \citet{Cheng_icuARM} investigated 40,000 readings collected from more than 30,000 ICU patients to provide LOS prediction as a decision support tool for the ICU physicians. \citet{basic2009admission} identified older patients who are suitable for admission to medical short stay units by determining the predictors of LOS. They used multivariable logistic regression to identify predictors of hospital LOS being three days or less. \citet{dinh2013deriving} used step-wise logistic regression to predict short stay admissions for trauma patients who are admitted to a major trauma centre. \citet{hachesu2013use} employed artificial neural networks, support vector machines and decision trees to predict LOS for patients with coronary artery disease. 

\citet{capuano2015factors} studied LOS in ambulatory patients arriving at an ED by excluding the admitted patients. Their dataset included attributes such as age, gender, day and month of the visit, and total number of annual emergency room visits. Their analysis indicated that only three of these attributes are significantly associated with LOS: age, clinical classification class of the patient and total number of ED visits. In a recent study, \citet{street2018influences} proposed a predictive model to identify the patients with an ED LOS of more than four hours. They developed a multivariate logistic regression model by using variables such as triage category, arrival overnight, arrival by ambulance, laboratory investigations, and time to be seen by a doctor. \citet{Cui2018} considered multi-output regression models to jointly predict cost and LOS associated with patients to better estimate the resource utilization in the hospitals. Their analyses were based on hospital inpatient records, which contain information on diagnosis and surgical operations. 

Some other studies in the literature focused on predicting LOS for particular patient types such as ICU patients \citep{houthooft2015predictive} or pneumonia patients \citep{khajehali2017extract}. Limiting the patients to certain categories in LOS prediction problems helps reduce sparsity and patient-level variation in LOS. Therefore studies with more limited patient types generally reported higher prediction performance values. For instance a study performed with cardiac patients reported an AUC value of 0.90 for the proposed prediction model \citep{rowan2007use}.

In addition to machine learning based methods for predicting the LOS at the time of admission, several other approaches have been considered to estimate the LOS for in-patients~\citep{baek2018analysis}. \citet{Papi2016} characterised the LOS distribution in an hospital using phase-type distributions. Similarly, \citet{Gordon2018} proposed a model based on Coxian phase-type distributions to predict the LOS for elderly patients. \citet{Vasilakis2005} compared various statistical techniques to model in-patient LOS for stroke patients using survival analysis, mixed-exponential and phase type distributions and compartmental models. \citet{pape2020predicting} used parametric and semi-parametric survival models to obtain the discharge probabilities of individual patients, which are then used to estimate the bed occupancy levels at the hospital.

The preponderance of evidence in this area is focused on predicting hospital LOS using a narrow set of demographic variables using conventional regression modelling approaches. Published models have varying degrees of predictive accuracy, with AUC values ranging between 0.68-0.90. Many of the inputs for traditional models may not be available in real-time electronic health records and those could not be used for a decision support system \citep{lu2015systematic}. We contribute to the current literature by using granular clinical data that are routinely available in electronic health records, including laboratory test results and use of diagnostic imaging, and by applying various machine learning approaches. Moreover, we develop a discrete-event simulation model that mimics the hospital operations with regards to allocation of the patients to short stay units or inpatient beds. The proposed simulation model offers insights on the potential impacts of using prediction models in clinical practice.

\section{Methods}\label{sec:methods}
Our aim in this study is to first predict the LOS using machine learning-based approaches, and then evaluate the impact of these predictions using a simulation model that mimics the patient flow in a hospital setting. 
Accordingly, we start by describing how patients are currently assigned to ward beds in a hospital setting. 
Afterwards, we review the machine learning models that are used for the prediction task, and introduce our simulation model.

\subsection{Hospital and ED Setting}
We base our analysis on a single hospital located in downtown Toronto. 
At present, the study hospital does not have a short stay unit. 
The hospital's current practice with regards to admitting patients from ED to general wards can be described as follows. 
When ED patients require general ward admission, patients are assigned to an inpatient bed by a nurse or manager in the ED, in consultation with nurses in charge at relevant inpatient wards. 
If additional input is required, the hospital's Director of Care may be involved. 
Patients are preferentially assigned to wards based on the admitting service (e.g., cardiology patients are admitted to the cardiology ward) and no consideration is given to estimated hospital LOS.
If the ideal patient ward is fully occupied, patients may be assigned to an off-service ward, and in some cases, patient complexity and estimated hospital LOS are considered in an ad hoc fashion. 
No standardized approaches are used to assist with inpatient bed assignment from the ED based on hospital LOS.

\subsection{Prediction Models}
Our dataset contains a LOS value for each encounter, which allows us to use both regression and classification techniques to build a prediction model. 
In particular, a regression model can be used to predict numerical LOS values, and then the patients can be classified as long stay (LS) and short stay (SS) using a threshold value (e.g., 72 hours) over the predicted LOS. 
Alternatively, we can first divide the patients as LS and SS based on the threshold value and then perform a classification task over the dataset. 
Our preliminary analyses indicated that classification techniques performed better than regression-based approaches based on our dataset, and classifiers may also be more useful for clinical workflows. 
Accordingly, we focus on classification techniques in our study. 

\tcolB{There are a large number of machine learning models that can be used for binary classification tasks (e.g., LOS prediction).
Some of these models have simpler forms.
For instance, logistic regression model relies on the logistic (sigmoid) function, that is, $\ell(\mathbf{x}) = \frac{1}{1+\exp^{-(w_0 + \mathbf{w}^{\intercal} \mathbf{x})}}$
where $w_0$ is the intercept, $\mathbf{w}$ represents the regression coefficients, and $\mathbf{x} = [x_1,\hdots,x_d]$ corresponds to the feature vector (e.g., lab results and age for LOS prediction) of a given data instance.
Once the intercept and coefficient values are estimated, the binary class prediction, $\hat{y}$, can be made based on the value of the $\ell(\mathbf{x})$ function, e.g., $\hat{y} = 1$ if $\ell(\mathbf{x}) \geq 0.5$, and $\hat{y} = 0$ otherwise.
For a logistic regression model, the importance of each feature/attribute can be assessed through intercept and coefficient values as well as other statistics such as odds ratio, which makes this model highly interpretable. }

\tcolB{Another classification model that we considered is Linear Discriminant Analysis (LDA) classifier~\citep{bishop2006pattern}, which uses discriminant functions in the form of simple probabilistic models, and generates the predictions for each instance $\mathbf{x}$ using the Bayes' rule, that is,
\begin{align}
\label{eq:lda_posterior} P(y=k | x) = \frac{P(x | y=k) P(y=k)}{P(x)}.
\end{align}
LDA classifier then returns class $k$ that maximizes the above posterior probability as the prediction.
Note that the term $P(x|y)$ in Eqn.~\eqref{eq:lda_posterior} is typically modeled as multivariate Gaussian distribution,
$\mathcal{N}_k(\boldsymbol{\mu}, \boldsymbol{\Sigma})$.
}

\tcolB{Decision Tree (DT) ensembles are also commonly used for binary classification tasks \citep{Breiman2001, Chen2016}.
DT models split the data repeatedly using certain cutoff values for the features, and through splitting, different subsets from the original dataset are created with each instance belonging to one particular subset.
DT model predicts the class labels as the average of the values for the instances in the leaf nodes (i.e., final subsets).
DT models are relatively simple, and intrinsically interpretable \citep{molnar2020interpretable}.
Feature importances in a DT can be computed by going over all the splits that involve the target feature, and measuring how much variance (or Gini index) is reduced compared to the parent node due to the split on the target feature.
Tree-based ensembles employ a number of DTs in a systematic manner to prevent overfitting and improve prediction accuracy.
Random Forest (RF) is a bagging ensemble that fits multiple DTs on various subsets of the dataset, and aggregate the individual predictions (e.g., by voting or averaging) to obtain the final prediction.
Gradient boosted trees and the variants such as XGBoost and Light Gradient Boosting Machines (LGBM) employ DTs as weak learners, and iteratively build a single strong learner.
Within a boosting scheme, these models add DTs to the ensemble one at a time, and fit them to correct the prediction errors that were made by the prior DT models.
While the tree-based ensembles are highly complex, since they are based on DTs, it is possible to obtain the feature importances for these ensemble methods as well.}

We also considered other classification methods such as multi-layer perceptrons, logistic regression (LogR), Na\"ive Bayes, $k$-nearest neighbor (KNN), support vector machines (SVM) and Gaussian process classifier. 
We provide a comparative analysis between different machine learning models trained for LOS prediction in Section~\ref{sec:MLModelResults}.


\subsection{Simulation Model for Patient Flow}
\begin{figure}[!ht]
    \centering
\includegraphics[scale=0.375]{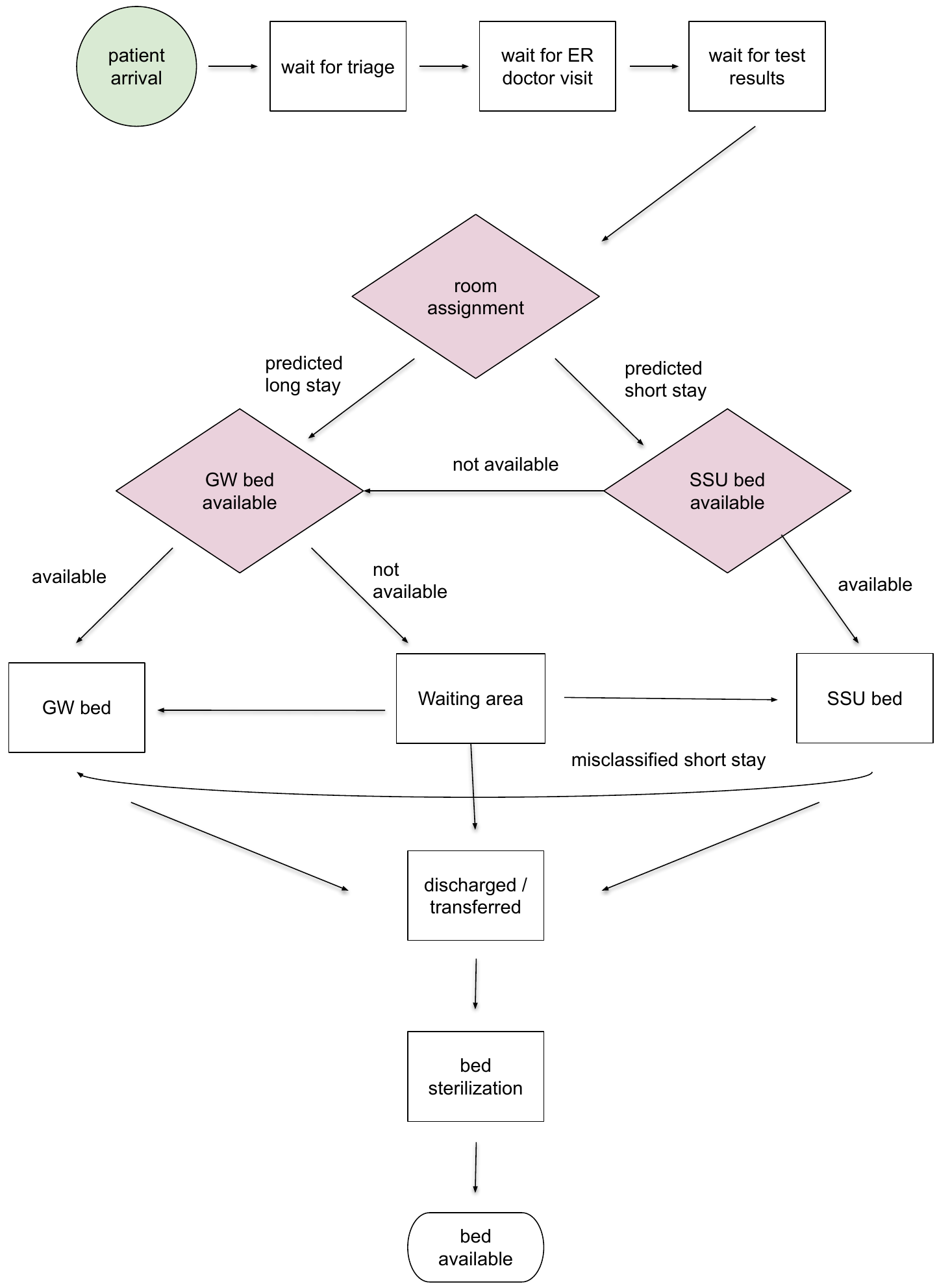}
    \caption{Flowchart of the proposed simulation model showing the patient movement in the hospital}
    \label{fig:simModel}
\end{figure}

A flow chart for our discrete-event system simulation model which mimics the patient flow between the ED and GIM is provided in Figure~\ref{fig:simModel}. 
In the simulation model, patients arrive according to their actual arrival times to the hospital, and go through a series of processes in the ED such as seeing the triage nurse and ED doctor, and getting the recommended lab/radiology tests. 
Then, based on the collected information, the patients are classified as an LS or an SS patient. 
LS patients are transferred to the general ward (GW), whereas SS patients are placed in the short stay unit (SSU). 
If there are no available beds, then the LS and SS patients wait in the waiting area. 
LS patients who are misclassified as SS patients are moved to GW after 72 hours of their stay. 
In the simulation model, we assume that the detection of misclassification for the LS patients occurs after they stay 72 hours in the SSU, however, in the clinical setting such misclassification detection may occur at any point in time. 
We also assume that misclassified SS patients are not transferred back to SSU and stay in the GW until their discharge. 
After a patient is discharged (or transferred from one unit to another -- e.g., from SSU to GW), the room is considered unavailable until sterilization process is complete.
Sterilization process duration is assumed to be uniformly distributed between 1-3 hours.

For the simulation analysis, we separated our patient dataset into training and test sets based on the arrival time of the patients to the ED.
Specifically, the training set only included patients who arrived between April 1, 2010 and December 31, 2013, and the test set only included the arrivals between January 1, 2014 and March 31, 2015. 
We used the training part of the dataset to train the LOS prediction models and the testing part of the dataset to evaluate the prediction models via simulation.

\section{Data Description and Preprocessing}\label{sec:data}
In this study, we use a subset of data that was collected by the GEMINI project \citep{verma2017patient}. 
Our dataset consists of 16,222 hospital encounters, which includes all emergency admissions to GIM at a single hospital between April 1, 2010 and March 31, 2015. 
GIM patients account for up to 40\% of admissions to hospital from the ED and represent the largest single group of inpatients \citep{verma2017patient}. 
The dataset contains 1,277 different features, and includes information about patient admission records (15), laboratory test results (1,182), radiology tests (44), and vital signs and clinical documentation (36). 
Specifically, admission records include information on demographics (e.g., gender) and circumstances of the admission (e.g., admission time, admitting unit, and whether the patient is admitted via ambulance). 
Laboratory and radiology tests include all of the tests sent for each patient. 
Vital signs and clinical documentation include recordings taken by the triage nurse, such as blood pressure and pain intensity levels. 
Importantly, for prediction, we include only values that are available prior to the time of hospital admission (i.e., data generated in the ED prior to the decision to admit the patient to hospital). 

\begin{figure}[!ht]
    \centering
    \includegraphics[scale=0.60]{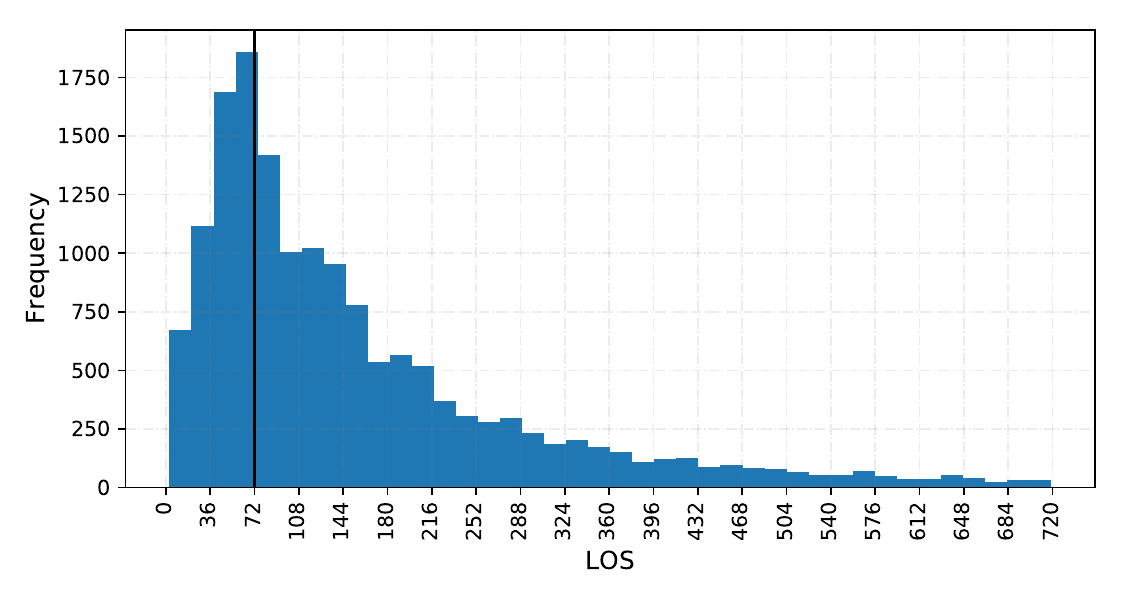}
    \caption{Distribution of LOS values in hours}
    \label{fig:LOS}
\end{figure}

Hospital LOS is calculated for each admission (the difference between discharge date and time and admission date and time), and this is used to label each admission as SS or LS using a threshold value of 72 hours. 
This value is selected because 72 hours is a time threshold commonly used for short stay units \citep{verma2019characteristics}. 
It is important to note that, based on the chosen threshold value, the dataset is imbalanced with 67\% of the records being LS and remaining 33\% being SS (see Figure~\ref{fig:LOS} for the distribution of LOS values). 
Furthermore, the dataset includes a high number of missing values considering that not all tests are performed on all patients and not all medical historical data is available for every patient. 
Typically, features with high missing ratios tend to provide little information on the patient category. 
Therefore, we removed the features with high missing ratios from our dataset by using a threshold missing ratio value (e.g., remove the features with a missing ratio of 90\%), given that the feature was not informative on the class label.
All the prediction models were tested with sparse (i.e., without imputing for missing values) and mean-imputed versions of the dataset (i.e., replacing the missing values with mean and mode of the feature values for numeric and categorical features, respectively), however, no significant performance differences between these two approaches were observed in our analysis.
In order to simplify the presentation of the results, only results with the imputed dataset are presented.

\subsection{Feature Engineering}
Identifying useful features within a given dataset is one of the most important building blocks of successful machine learning applications. 
As in other domains, there are several studies focusing on feature engineering within a specific type of health care dataset (e.g., see \citep{garali_feature_selection}, \citep{nabian_feature_selection}, and \citep{Yadav_BurnDiagnosis}). 
In our analysis, we observed prediction performance improvement by using additional features generated by applying basic statistical analysis over the available data. 
These generated features were also found to be important features by the feature selection methods described in the following section, providing evidence for the validity of the engineered features.

In our feature engineering effort, we started with simple descriptive analysis-based features such as day of presentation to hospital
(triage\_dayofweek), the time patients wait between triage and hospital admission (wait\_time\_to\_admit), and number of vital sign measurements collected in the ED portion of the encounter (vit\_current\_visit\_vital\_test\_count). 
Next, we focused on identifying patient-specific information. 
Since each piece of data is identified with an encounter identification string and encounter ids were matched with a hash code, we are able to use some of the information from past ED visits of individual patients, where available. 
This approach led to the introduction of the following features: number of encounters for the same patient (previous\_encounter\_count), hospital LOS in hours of the most recent previous visit of the patient (previous\_encounter \_most\_recent\_los), number of lab tests performed in all previous visits  (previous\_lab\_test\_count), number of unique lab tests performed in all previous visits of the patient (previous\_lab\_test\_uniq\_count), and number of the most frequent diagnosis codes assigned to the patient in previous visits (previous\_ip\_diag\_count). 
We also introduced features that are obtained through simple data manipulations such as number of lab tests with results higher/lower than the lab test result at the 75th percentile (outlier\_lab\_result\_with\_percentile\_75), and number of total, high and low blood pressure readings (vit\_total / high / low\_bp\_count).
Lastly, we considered one-hot-encoding for categorical features such as ``triage\_dayofweek'' and ``admitting\_service''. 

\subsection{Exploratory Data Analysis}
\begin{figure}[!ht]
	\centering
	\subfloat[\textbf{Triage time} \label{fig:dataTriageTime}]{\includegraphics[scale=0.45]{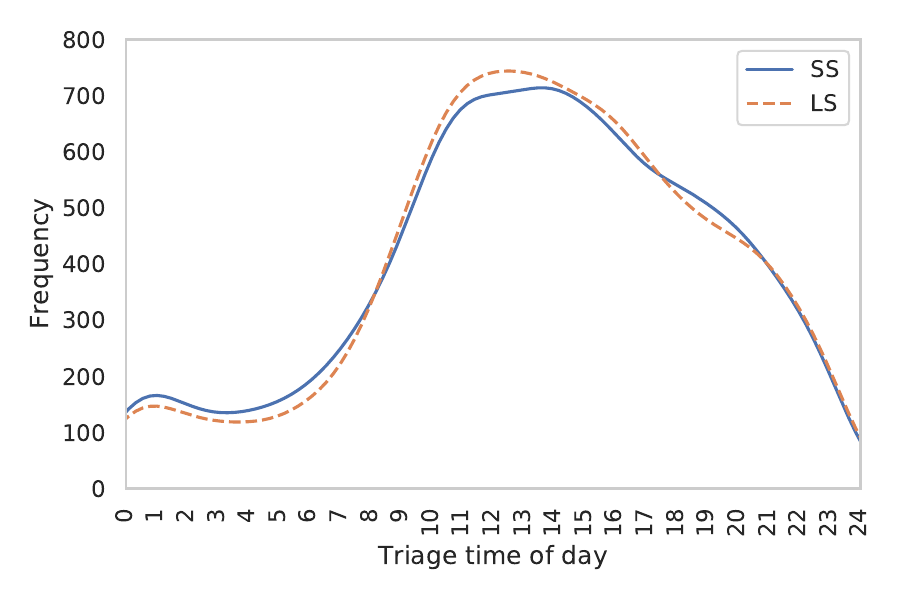}}
	\subfloat[\textbf{Admissions} \label{fig:dataAdmissions}]{\includegraphics[scale=0.45]{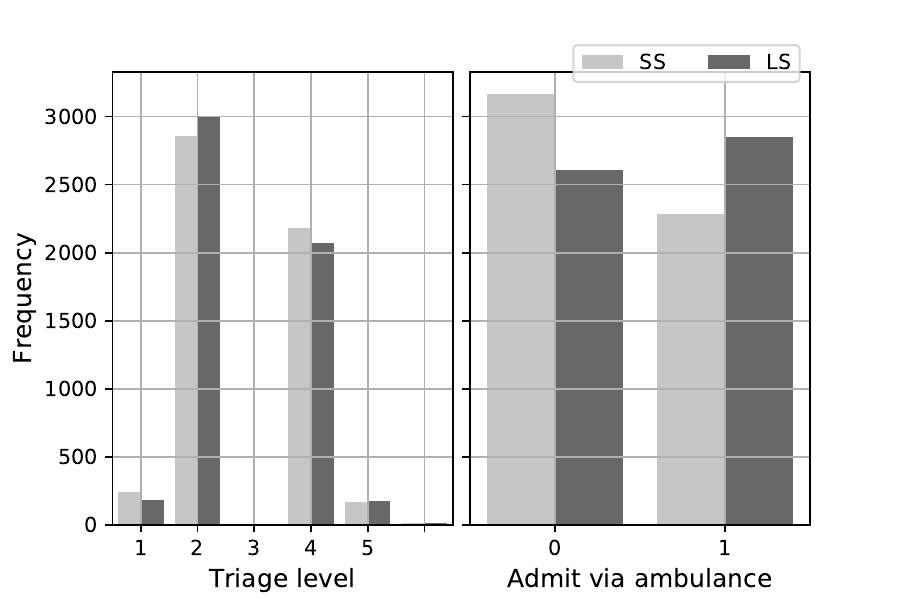}} 
	\par
	\subfloat[\textbf{Lab test count vs. age} \label{fig:dataTestCtAge}]{\includegraphics[scale=0.45]{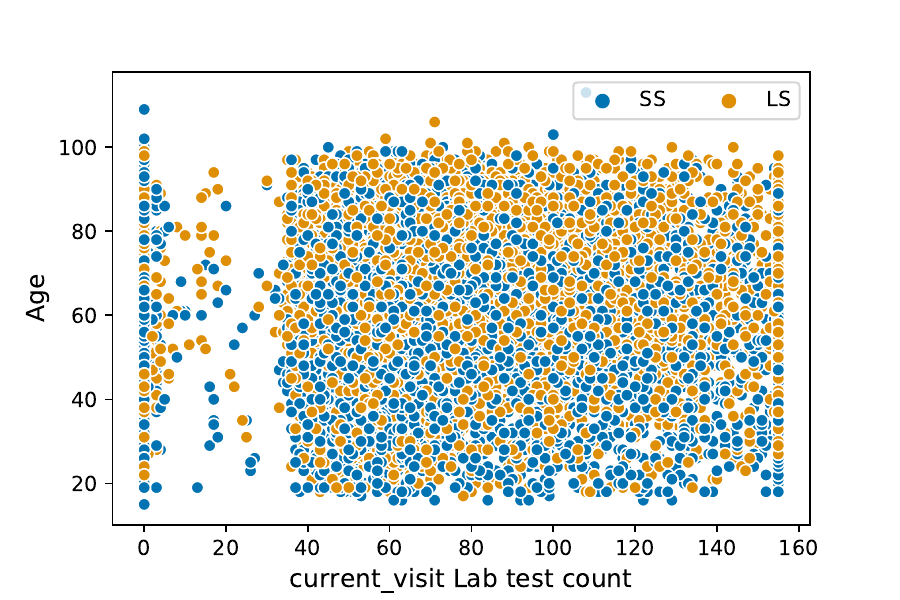}} 
	\subfloat[\textbf{Abnormal lab results} \label{fig:dataBadLabRess}]{\includegraphics[scale=0.45]{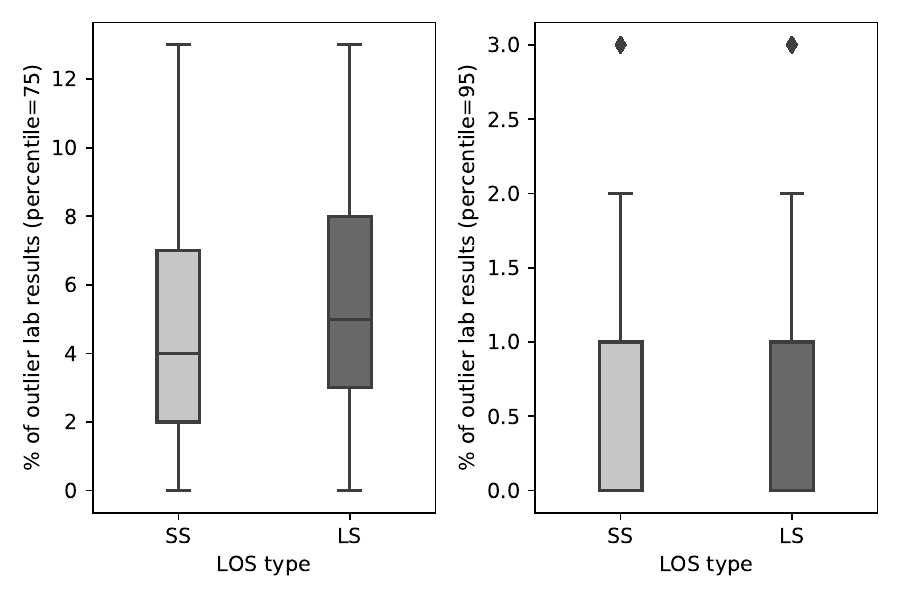}} 
	\caption{Observed trends for various attributes in the dataset} 
	\label{fig:data}
\end{figure}

Depending on the types of attributes under consideration, we conducted appropriate univariate and bivariate analysis for each attribute in order to better understand the overall structure of the dataset and associations with hospital LOS. 
For categorical variables, we measured the frequency of each category, whereas for numerical variables we inspected variation (e.g., variance, standard deviation and interquartile range), and central tendency (e.g., mean, median and mode). 
Figure~\ref{fig:data} summarizes sample findings through exploratory data analysis. 
Note that, we used a balanced dataset (with identical number of SS and LS patients) to plot this figure.

Figure~\ref{fig:dataTriageTime} shows that the proportion of LS and SS patients does not appear to vary based on triage time of day. 
In Figure~\ref{fig:dataAdmissions}, we observe that triage acuity measured by the Canadian Triage and Acuity Scale (levels 1-5, with 1 being the most emergent case) assigned by the triage nurse does not seem to be helpful in identifying the LOS class. 
Similarly, from Figure~\ref{fig:dataTestCtAge}, we observe that correlation between patient age and number of lab tests performed on patient is not a good predictor for LOS. 
Finally, Figure~\ref{fig:dataBadLabRess} demonstrates that the number of ``outlier'' lab results that fall outside of 75 percentile value may be informative as LS patients tend to have a higher number of outlier lab results on average.

\subsection{Feature Selection}\label{sec:MethodsFS}
We considered common feature selection (FS) techniques to assess the importance of each attribute in the dataset. 
We first analyzed the correlation between the attributes. 
For this purpose, we started by dividing the features into sets of numerical and categorical features, and employed Spearman's and Pearson's correlations where appropriate (e.g., Pearson for numerical features and Spearman for ordinal features). 
In order to remove redundant attributes, we only kept one of the features where correlation of two input variables were larger than some predetermined value (e.g., 0.9). 
In addition, we conducted bivariate analysis for feature selection. 
\tcolB{We employed several statistical tests such as chi-square test and F-statistic as well as information gain approaches to analyze the relation between individual features and the class label. 
Moreover, we combined the FS methods that are used for bivariate analysis, namely, Pearson correlation, Spearman correlation, Cramer's V measure, chi-square test, F-statistic, and mutual information to come up with an ``Ensemble FS'' technique.
Specifically, Ensemble FS ranks the features according to an aggregate score determined by their rankings based on the individual FS techniques.}

\section{Results}\label{sec:results}
We conduct a detailed numerical study to examine the effects of incorporating the LOS prediction 
in a hospital setting.
We first provide results on the performances of machine learning-based models for predicting LOS category of the patients. 
Then, we discuss the results of our simulation model which is used to evaluate the impact of the prediction models in practice.

\subsection{Prediction Model Results}\label{sec:MLModelResults}
We use the Scikit-learn Python library in our analysis \citep{Pedregosa2011}. 
\tcolB{The final recommendations produced by the machine learning algorithms were evaluated using accuracy, precision, recall and F1-score metrics, which can be defined as
\begin{small}
\begin{align*}
& \mbox{precision} = \frac{\text{TP}}{\text{TP} + \text{FP}} \quad && \mbox{recall} = \frac{\text{TP}}{\text{TP} + \text{FN}}\\[0.3em]
& \mbox{accuracy} = \frac{\text{TP} + \text{TN}}{\text{TP} + \text{TN} + \text{FP} + \text{FN}}\quad && \mbox{F1-score} = 2\cdot \frac{\mbox{precision} \cdot \mbox{recall}}{\mbox{precision} + \mbox{recall}}
\end{align*}
\end{small}
\noindent where TP, TN, FP, and FN refer to the number of true positive, true negative, false positive, and false negative classifications, respectively. 
Note that we use weighted F1-scores, which is the average F1-score calculated for each label obtained by taking into account the proportion of each label in the dataset.}

The importance of the performance metric typically depends on the application problem. 
For instance, in cancer diagnosis, false negatives (i.e., missing cancer) are particularly harmful, and hence recall can be considered to be more important than precision. 
\tcolB{For the LOS prediction problem, we note that as the LS class is assumed to be the positive class, we consider recall and precision of the LS class to be the generic recall and precision metrics for classification.
One particular objective in predicting whether the patients are LS or SS is to correctly identify SS patients to assign those to SSUs. 
Due to relatively small capacity of SSUs compared to GWs, recall (i.e., sensitivity) could be deemed to be more important than precision (i.e., positive predictive value). 
A high recall value implies a lower number of LS patients being misclassified as SS (e.g., through fewer FN instances), hence preventing those to occupy SSU beds.
In addition, low recall values imply frequently labeling the LS patients as SS, which might lower the quality of care for the LS patients when they are placed in SSUs. 
In this regard, for the sake of being more selective about a particular performance metric such as recall, 
a model calibration can be performed, e.g., by determining the precision at a specific recall threshold (e.g., 90\%).
On the other hand, misclassification of SS patients may also be undesirable since they would occupy GW beds, and might increase the GW bed waiting time for LS patients.
Accordingly, when comparing the prediction models, 
we focus on the F1-scores, which balance precision and recall values.}

\tcolB{We followed the data pre-processing steps discussed in Section~\ref{sec:data} to prepare the dataset for machine learning model training.
For any feature pair with a correlation value higher than $\rho = 0.99$, we arbitrarily discarded one of the features from the dataset.
This threshold is set to a very high value to prevent any information loss.
For instance, while blood hematocrit and hemoglobin counts are highly correlated ($\rho = 0.978$), they are both used to test for anemia.
Regardless, this feature elimination procedure helped eliminating a few features from the dataset (e.g., engineered feature ``radiology unique test count'' is discarded due to having almost perfect correlation with another engineered feature ``radiology test count'').
In an attempt to reduce the sparsity level of the dataset, we eliminated the features that are infrequently observed in the dataset by removing the features with 75\% or higher missing ratios.
We created new features from the removed features to extract some useful information.
For instance, for the removed lab tests, we added a new feature named ``lab test count rare'', which shows how many of these removed lab tests are done for each patient.
These operations resulted in a set of 114 selected features out of 1,277 available features, which includes 78 lab test-related features, 7 radiology test-related features, 12 vital signs-related features, and 17 admission record-related features.
We also experimented with different thresholds for the missing ratios.
Setting the threshold to 90\% led to 152 features, however, our preliminary analysis did not reveal any prediction performance improvement due to having the additional features with higher missing ratios in the dataset.}

\subsubsection{Comparison of the prediction models}\label{sec:compareML}
We compare a wide variety of machine learning models for predicting how long GIM patients stay in the hospital.
\tcolB{We consider both undersampling and oversampling strategies to cure the data imbalance problem.
We used the \textit{SMOTE} function from \textit{imblearn} library in python for data balancing.
Because SS is the minority class, we randomly downsampled the LS instances for undersampling, and oversampled the SS instances for oversampling, which led to 10,906 instances for the former, and 21,538 for the latter.
We apply 10-fold cross-validation to show how the prediction models generalize to the previously unseen test instances~\citep{wainer2018nested}.}
Note that our experiments with the grid search approaches to tune the parameters for the machine learning models indicate that there is no significant performance gains due to the tuned parameters. 

\begin{figure}[!ht]
	\centering
	\subfloat[Undersampling LS instances (avg. F1-scores -- DT: 0.548, LogR: 0.621, RF: 0.630, LDA: 0.634, LGBM: 0.638
)]{ \includegraphics[width=0.93\textwidth]{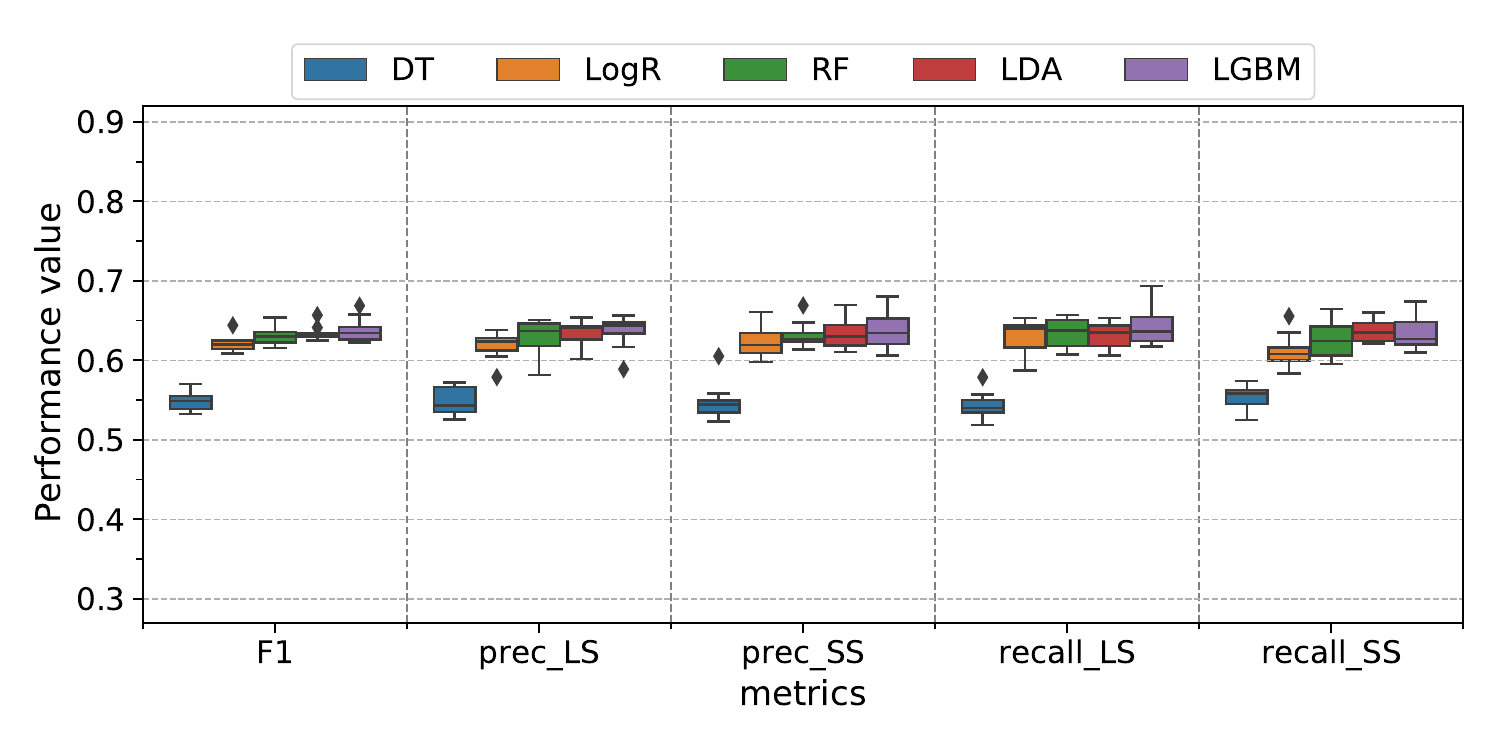}
	}\par
	\subfloat[Oversampling SS instances (avg. F1-scores -- DT: 0.592, LogR: 0.629, RF: 0.657, LDA: 0.659, LGBM: 0.664
)]{
	\includegraphics[width=0.93\textwidth]{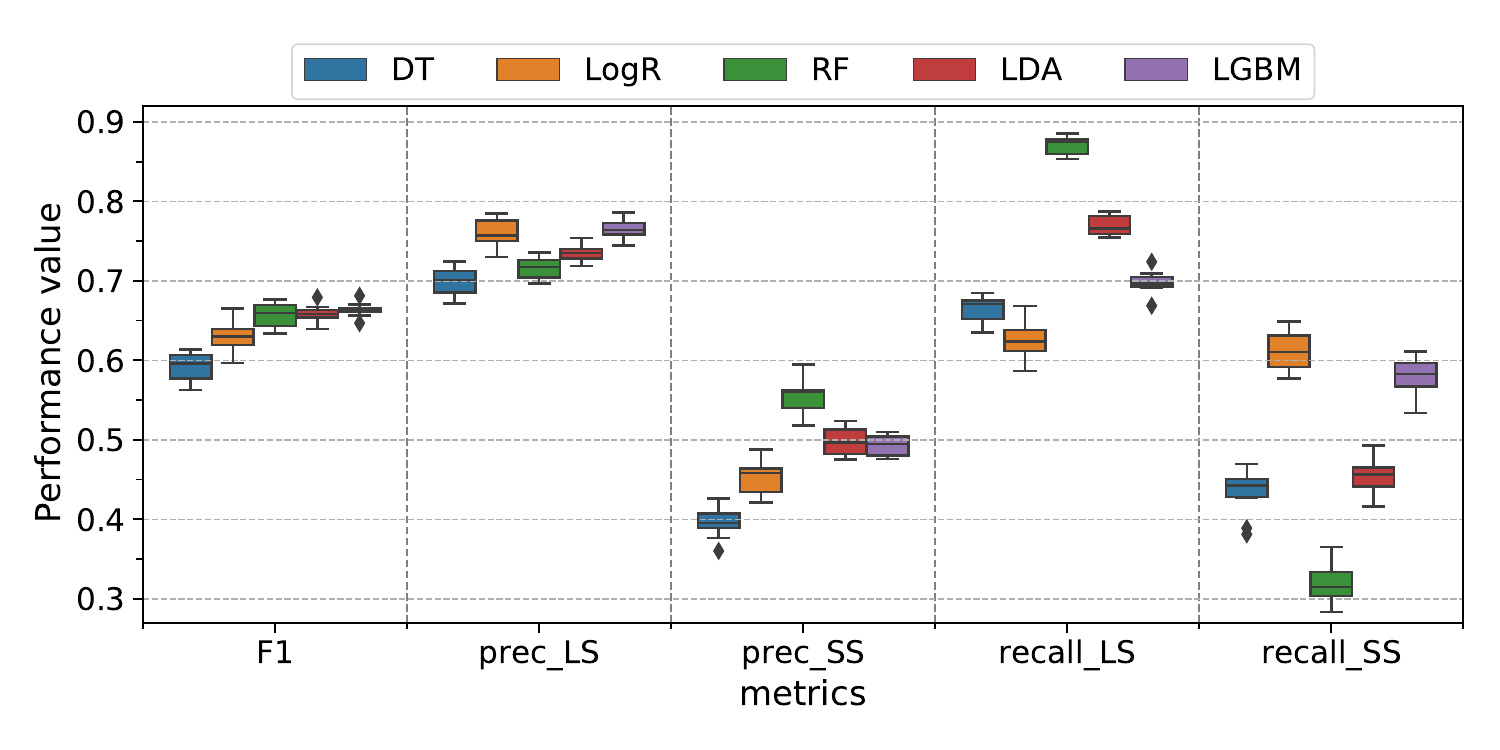}}
	\caption{\tcolB{Comparative results for different machine learning models using 10-fold cross-validation}}
	\label{fig:compML}
\end{figure}

Figure~\ref{fig:compML} shows the summary results from our experiments with different prediction models, namely, DT, LogR, RF, LDA, and LGBM using 10-fold cross-validation.
\tcolB{We observe that the relative performances of the prediction models were not significantly affected by the data balancing strategy. 
Specifically, based on the F1-scores (F1), the performances of RF, LDA and LGBM models are relatively close to each other, and DT performs the worst.
We also note that the overall performance of the prediction models are slightly improved by oversampling, as indicated by the F1-scores.
While undersampling provides a more uniform performance for predicting SS and LS instances, oversampling increases the prediction performance for LS instances at the expense of reduced precision and recall values for SS instances.
This result might be attributed to the fact that while SS instances are synthetically generated in the case of oversampling, the models are introduced a more diverse set of actual LS instances, which might make them more likely to predict instances as LS. 
For oversampling, we find that LGBM and LDA provide a relatively more uniform performance for predicting SS and LS instances, and RF performs poorly for correctly classifying the SS instances as indicated by low recall\_SS values.
Lastly, we note that as the LS class is assumed to be the positive class, recall\_LS and precision\_LS correspond to the generic recall and precision metrics for classification, respectively.}

\tcolB{We also consider statistical tests to compare the performances of the prediction models.
We use $5\times 2$-fold cross-validation paired t-test proposed by \citet{dietterich1998approximate} to perform pairwise comparisons.
At the significance level of $\alpha=0.05$, we find that, when comparing LGBM with LDA and RF over F1-scores, we fail to reject the null hypothesis (with $p$-values of 0.72 and 0.85, respectively, for LDA and RF), implying that observed differences in model performance are not significant.
On the other hand, in terms of F1-scores, LGBM performance is found to be different than LogR and DT  (with $p$-values of $10^{-5}$ for both LogR and DT).}

It is important to note that we considered other classification methods as well, namely multi-layer perceptrons, Naive Bayes, KNN, SVM, XGBoost, and Gaussian process classifier among others, and the presented comparison is not done on an exhaustive list of classification methods. 
In presenting the results, we specifically focused on the prediction models from different families that show relatively high predictive performance.



\begin{figure}[!ht]
	\centering
	\subfloat[Precision-recall curve]{\includegraphics[width=0.475\textwidth]{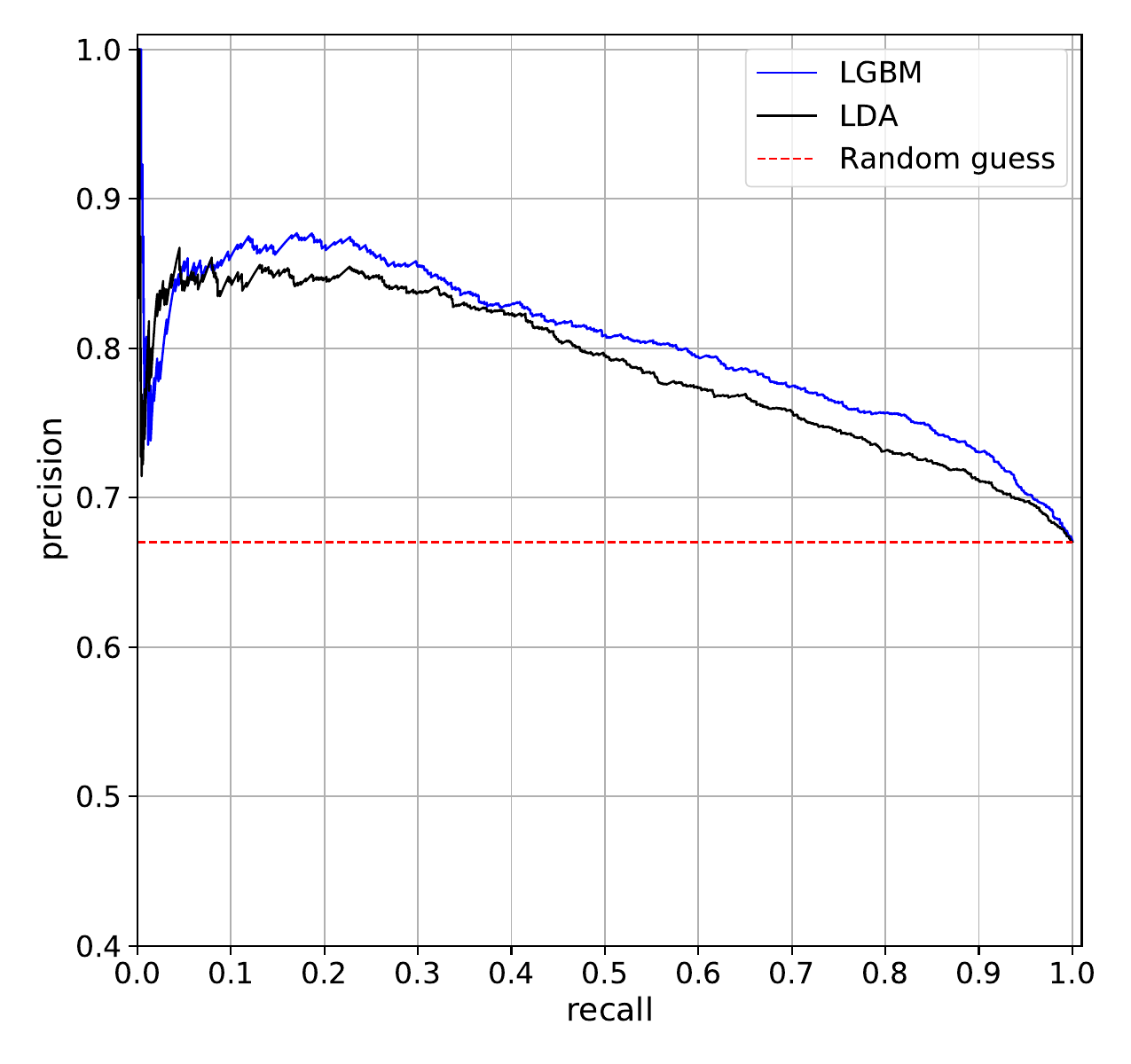}}
	\subfloat[ROC curve]{\includegraphics[width=0.475\textwidth]{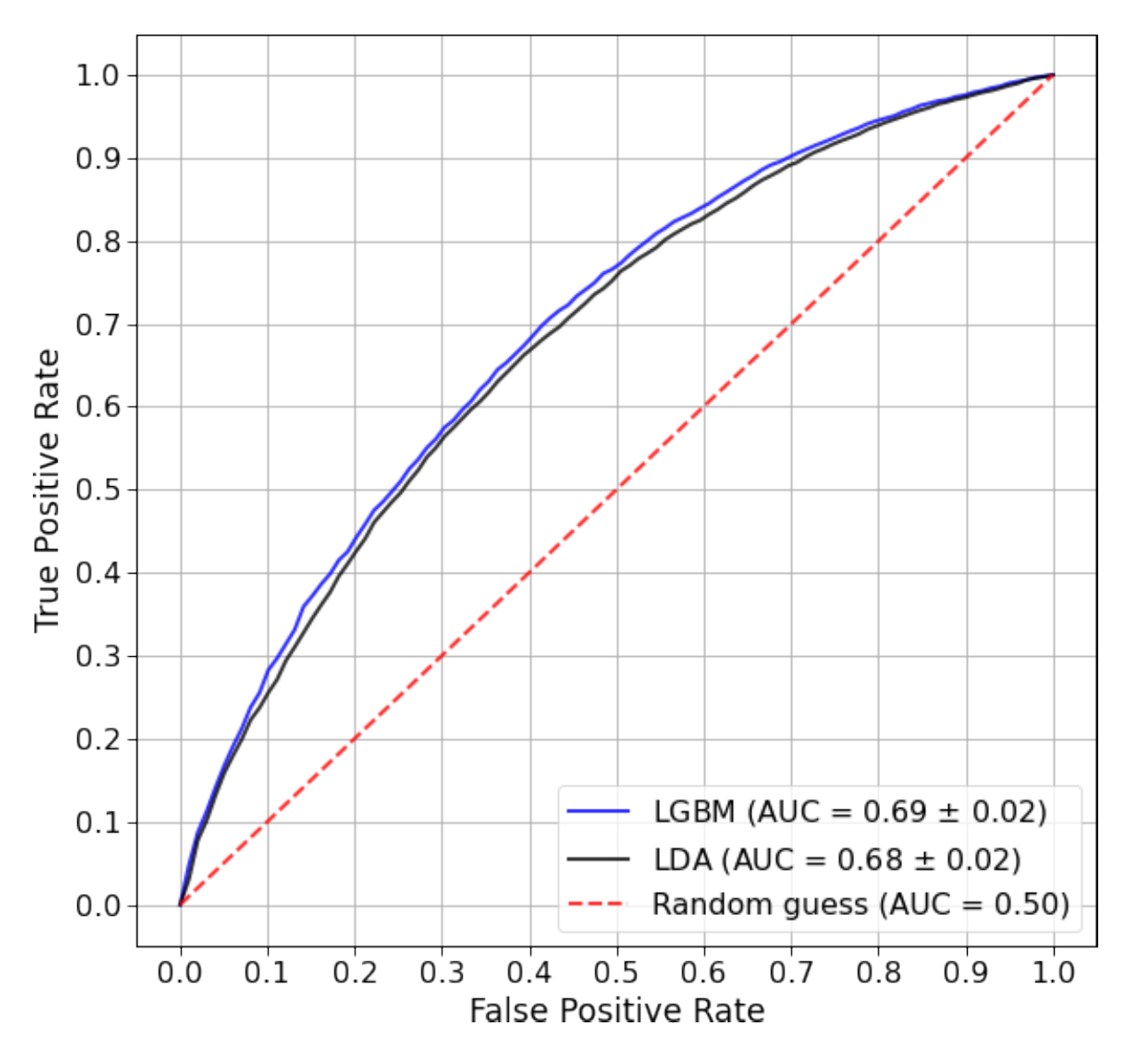}}
	\caption{Precision-recall and ROC curves for LDA and LGBM models (plotted using 10-fold cross-validation)}
	\label{fig:roc}
\end{figure}

In the remaining experiments, we use LDA and LGBM for LOS prediction, as they show the best performance among the considered prediction models.
In addition, we only consider oversampling when training the models.
Figure~\ref{fig:roc} shows precision-recall and receiver operating characteristic (ROC) curves, respectively, for LDA and LGBM models, which are obtained by 10 fold cross-validation (e.g., average of 10 ROC curves is plotted).
Both visualizations demonstrate that LGBM slightly overperforms LDA at various performance levels.
Area under curve (AUC) value, 0.69, is also slightly higher for LGBM, compared to LDA with an AUC value of 0.68, while both have similar standard deviation (0.02) for AUC.
\tcolB{We also use permutation testing to make pair-wise comparisons over ROC curves of different machine learning models \citep{venkatraman2000permutation}.
We find that AUC differences between LGBM and two other best performing models, LDA and RF, are statistically significant at significance level of $\alpha=0.05$, as indicated by the $p$-values of 0.001 and 0, respectively.
On the other hand, AUC difference between LDA and RF is not statistically significant ($p$-value of 0.166).}

\tcolB{Both precision-recall and ROC curves can be used to understand the trade-offs in model performances towards particular classes.
We can balance the precision, recall (i.e., TP rate) and specificity (i.e., 1 - FP rate) values based on the problem characteristics.
Specifically, we construct precision-recall curves over the training set to identify different prediction probability threshold values that can be used select the best LGBM model at certain achievable precision (e.g., 0.800) and recall (e.g., 0.900) values.
These threshold values are found to be 0.41 and 0.65, which lead to precision/recall values of 0.725/0.90 and 0.80/0.53, respectively.
Accordingly, we identify two new configurations for the LGBM model: high recall variant (LGBM\_r) and high precision variant (LGBM\_p).
As expected, a lower threshold value for the positive class label (i.e., LS), would lead to an increase in TP and FP predictions, and result in higher recall and lower precision values.
Alternatively, a high threshold value implies lower number of TP and FP predictions, resulting in higher precision and lower recall values.}


We next assess the performance of the prediction models by using a train-test split.
That is, the models are trained on the training set, and evaluated on the test set.
The dataset is split based on the time index, and the instances associated with patients arriving before January 1, 2014 are used for training, and the rest is used for testing.
Note that the same test set is used for the simulation experiments as well.
Accordingly, this experiment provides a more direct performance evaluation of the prediction models for simulation purposes compared to the previous experiments that employed cross-validation.
Table~\ref{tab:confusion} displays the number of TP, TN, FP and FN classifications, as well as mean, median and standard deviation of actual LOS values for the samples falling into TP, TN, FP and FN predictions of the LGBM model. 
\tcolB{As expected, compared to the baseline LGBM model, LGBM\_r (i.e., high recall variant) leads to a substantial increase in the number of positive (LS) predictions (TP+FP).
On the other hand, LGBM\_p (i.e., high precision variant) predicts more patients to be SS (i.e., negative class).
Accordingly, depending on the SSU capacity, LGBM\_p or LGBM\_r might be more preferrable over LGBM.}
The samples falling into the TP predictions category (i.e., LS patients classified correctly) with LGBM have an actual mean LOS value of 262.8 hours with a standard deviation of 309.6 hours. 
Similarly the samples in FN category (i.e., LS patients classified as SS) have an actual mean LOS value of 176 hours with a standard deviation of 168.1 hours. 
These findings indicate that LOS values for LS patients have a very wide range, once more demonstrating the difficulty of discriminating between SS and LS patients within the dataset.
On the other hand, the mean LOS value for patients with falsely predicted SS (i.e., FN) was still substantially shorter than patients with true LS predictions (mean LOS 176 vs 262.8 hours), suggesting that models offer meaningful information even when predictions are ``incorrect''. 

\setlength{\tabcolsep}{6pt}
\begin{table}[!ht]
    \centering
    \caption{Results on a test set of size 4,597 instances with 3,040 LS (i.e., positive) and 1,557 SS (i.e., negative) samples broken down into TP, TN, FP, and FN  predictions.}
    \scalebox{0.85}{
    \input{tables/MLRess_revision2/confusion.tex}
    }
    \label{tab:confusion}
\end{table}

\subsubsection{Impact of feature selection}\label{sec:impactOfFS}
\begin{figure}[!h]
	\vspace{-0.5em}
	\centering
	\subfloat[\scriptsize{Ensemble FS} \label{fig:feature_importance_pearson}]{\includegraphics[width=0.33\textwidth]{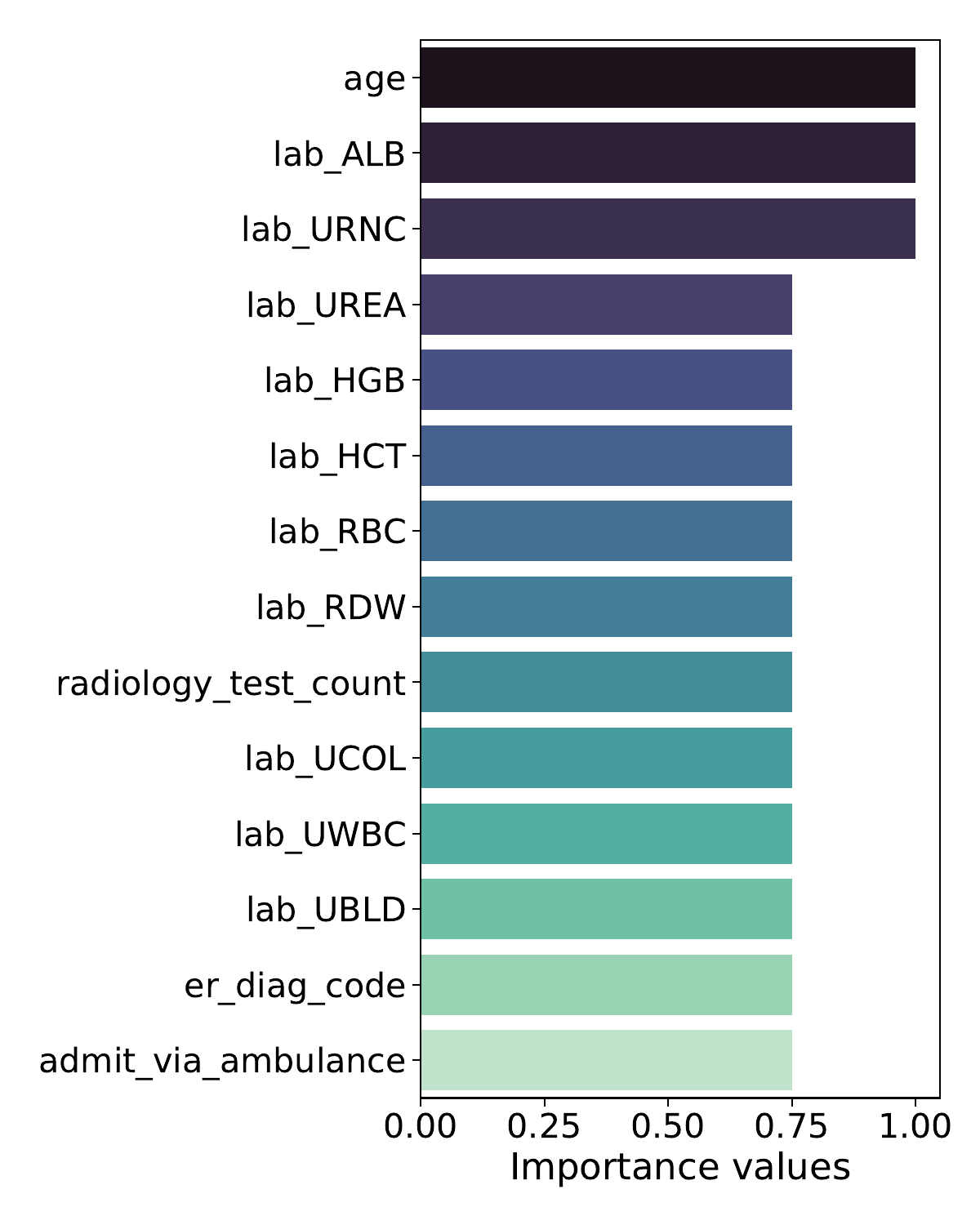}}\hfill
	\subfloat[\scriptsize{RF - feature\_importances} \label{fig:feature_importance_RF}]{\includegraphics[width=0.33\textwidth]{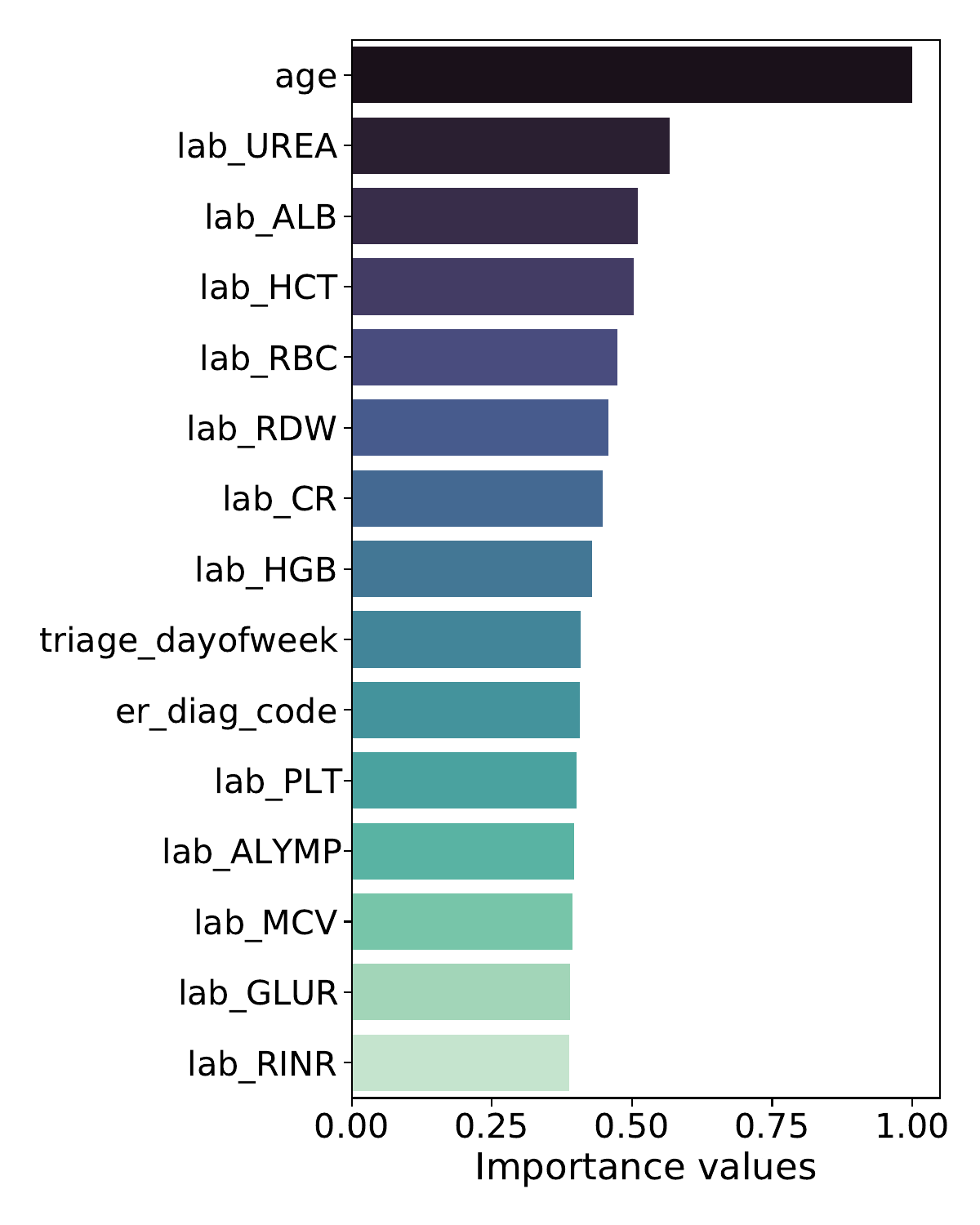}}\hfill 
	\subfloat[\scriptsize{LGBM - feature\_importances} \label{fig:feature_importance_LGBM}]{\includegraphics[width=0.33\textwidth]{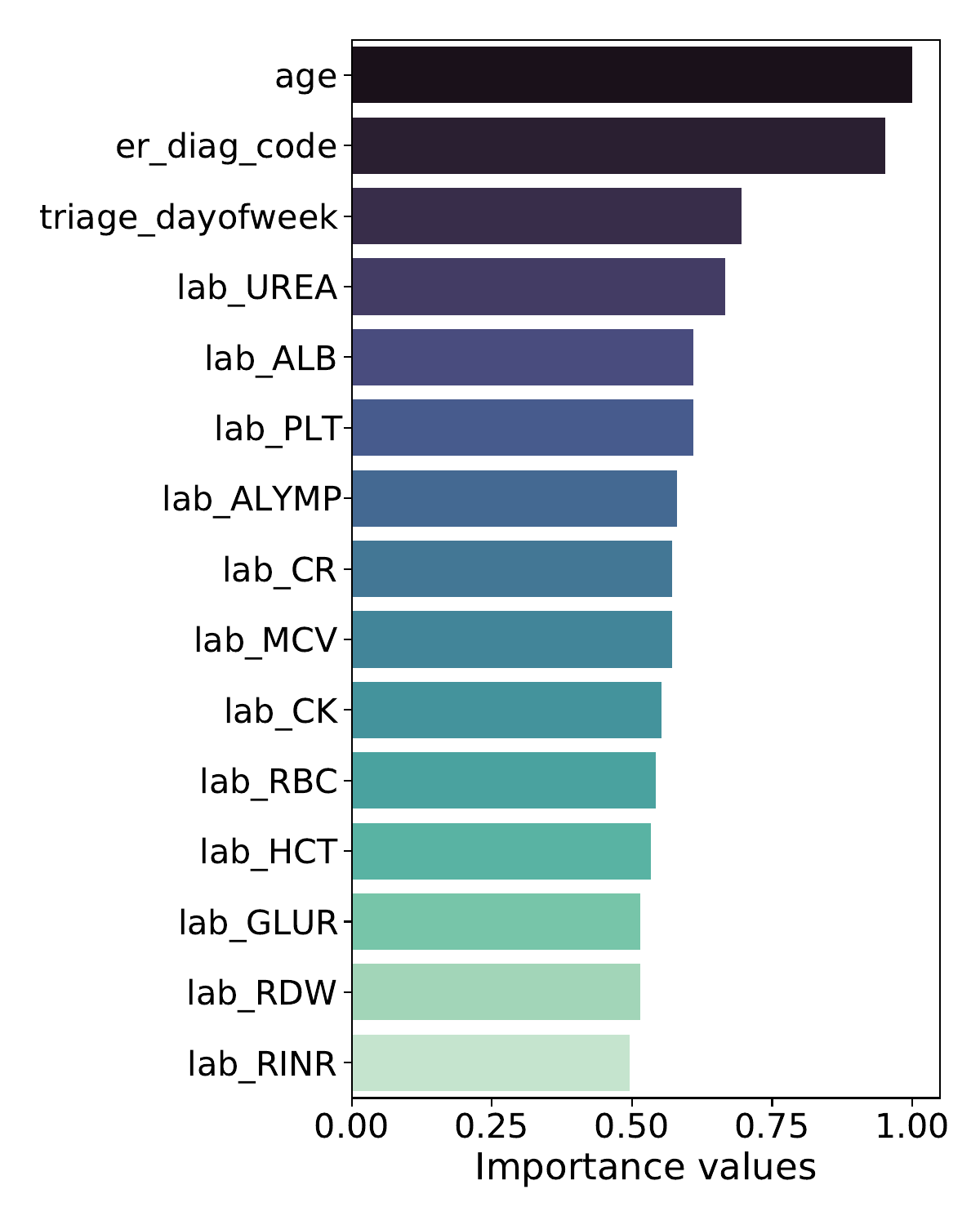}} 
	\caption{Relative importance values for the top 15 features obtained by Pearson correlation and through feature importance values from RF and XGB models} 
	\label{fig:feature_importance}
\end{figure}

We next investigate the impact of the individual features on the LOS classification. 
Our analysis with several feature selection strategies discussed in Section~\ref{sec:MethodsFS} did not reveal strong relations between any of the features and the class label.
\tcolB{For instance, two features with the highest correlation to the class label are age and lab\_UREA, which have Pearson correlation values of 0.188 and 0.121, respectively.
We provide relative feature importance values (normalized between 0 and 1) for the top 15 features obtained by Ensemble FS (see Section~\ref{sec:MethodsFS}), which is model agnostic, and feature\_importances method using RF and LGBM models in Figure~\ref{fig:feature_importance}. }

These results indicate that a few of the features stand out with slightly higher importance values in each case (e.g., age, lab\_UREA, and lab\_ALB).
However, different methods do not reach a consensus in the most important features.
Few other studies focused on identifying the important features linked to LOS at the time of admission. 
For instance, \citet{turgeman2017insights} used a dataset obtained from Veteran Health Administration hospitals, which contains information on patient demographics (e.g., age and gender), previously diagnosed comorbidities, inpatient and outpatient history, medication history, lab values (e.g., albumen and urea) and vital signs. 
Their analysis indicated that inpatient and outpatient histories were dominant predictors of LOS, which is expected considering the patient characteristics in the dataset. 
In another study, \citet{whellan2011predictors} found that, for heart failure patients, LOS increases with comorbidities and disease severity. 


We next report the impact of feature selection on predictive performance. 
\tcolB{In Table~\ref{tab:fs}, we present 10-fold cross-validation results for 40, 15, and 5 most important features obtained via ensemble FS and feature\_importances routine of the LGBM model.}
Note that LDA model does not provide an intrinsic method to assess feature importances, therefore we use the features identified by LGBM feature\_importances routine for LDA as well.
First, we note that, as also observed in Figure~\ref{fig:compML}, standard deviation for the F1-scores (F1-score std.) obtained from 10 fold cross-validation are low, indicating good generalizability of the prediction model performances for each case.
We find that feature\_importances routine is more successful in identifying the important features than the ensemble FS for LGBM, and provide comparable performance to ensemble FS for LDA.
We observe that, in the case of LGBM, using only the top 40 and 15 features (see Figure~\ref{fig:feature_importance_LGBM} for the list of 15 features) identified by feature\_importances causes relatively small decrease in F1-scores (0.6\% and 2.0\%, respectively), however using the top 5 features result in a more substantial deterioration in model performance (e.g., 5.1\% decrease in F1-score).
Note that feature selection are not likely to improve the prediction performance for many machine learning models.
For instance, tree-based methods (e.g., LGBM) have built-in feature selection capabilities, and can identify the most important features inherently.
However, these observations can be used to interpret the model predictions, e.g., to understand which features are used when classifying the individual instances.
This can be used to create a higher level of trust towards using these models in clinical practice.
Furthermore, they can guide future data collection processes in ED.
For instance, seeing that lab\_UREA and lab\_ALB are important for LOS classification, a particular care can be given to prevent wrong data entries for these tests.


\setlength{\tabcolsep}{4pt}
\renewcommand{\arraystretch}{1.25} 
\begin{table}[!h]
    \centering
    \caption{Impact of feature selection}
	\subfloat[LDA]{
	\begin{footnotesize}
    \scalebox{0.83}{
	\-\hspace{-0.99cm}\input{tables/MLRess_revision2/tbr_LDA.tex}
	}
	\end{footnotesize}
	}
	\\[0.5em]
	\subfloat[LGBM]{
	\begin{footnotesize}
    \scalebox{0.83}{
	\-\hspace{-0.99cm}\input{tables/MLRess_revision2/tbr_LGBM.tex}
	}
	\end{footnotesize}
	}
	\label{tab:fs}
\end{table}



\subsubsection{Discussion on prediction model results}\label{sec:results_discussion}
Our numerical study provides insights on several aspects of using machine learning-based predictive modeling for LOS prediction. 
Our first step in model development was data cleaning and pre-processing. 
In this step, we performed feature engineering to extract useful information from the raw data as well (e.g., identifying a feature showing the ``percentage of outlier lab results'' which was not immediately available in the dataset).
We also performed feature selection to understand the links between class labels and the individual features. 
For instance, we learned that results of certain lab tests such as albumin and urea are relatively highly correlated with the LOS values compared to other features in the dataset. 
Next, comparison of a wide range of machine learning models for LOS prediction showed that our dataset consisting of 16,222 instances with features obtained from patient admission demographics, laboratory test results, diagnostic imaging, vital signs and clinical documentation was not sufficient to predict LOS with high accuracy. 
Additionally, we assessed the prediction model sensitivity to training set size through increasing the number of instance in the training set incrementally (the results of this experiment were not provided for the sake of brevity).
We observed that after a certain number of training instances (e.g., 6,000), there is a diminishing return for further increase in the training set size.
\tcolB{These findings indicate that more informative features (and larger datasets) are needed for developing high-performing models for LOS prediction.}

\subsection{Simulation Results}
We evaluate different classification approaches using our simulation model: perfect classification, random classification, and the variants of LGBM. 
We do not consider LDA for the simulation analysis as LDA prediction performance is similar to LGBM\_p, and variants of LGBM provide a more direct opportunity to understand the impact of various performance metrics (e.g., precision and recall) for practical purposes. 
The random classification, which assigns an SS or LS label to each instance randomly, is considered as the baseline prediction model.
We provide four different groups of simulation outputs for each classification method: 1) number of bed sterilizations, 2) average GW bed wait time of patients, 3) number of misclassifications and 4) unit occupation statistics. 
The total number of times beds needed to be sterilized increases with an increased number of patient relocations, and so it can be used to quantify the impact of misclassifications in the system. 
Average waiting time for GW beds and average time spent by patient type in each unit can be used to evaluate the real-life impact of misclassifications.
In addition, GW and SSU occupation statistics such as average number of patients in GW/SSU (which are averaged over time) provide insights on utilization as well as congestion in these particular units.

We run our simulation model considering two configurations: infinite capacity hospital and finite capacity hospital. 
In the infinite capacity setting, which is taken as the baseline case, there is an infinite number of beds in both GW and SSU. 
In the finite capacity (i.e., `capacitated') case, we assumed 70 beds in GW, 10 beds in SSU.
We also provide results with the case where there are 70 beds in GW, no SSU (i.e., 0 capacity) to examine the impact of having SSU in the hospital.
Table~\ref{tab:simRessInfCap} shows that, in the case of infinite capacity, misclassifications can lead to substantial increase in total number of sterilizations, particularly due to wrongly assigning LS patients to SSUs as they are transferred to GW after 72 hours.
For instance, LGBM and random classification lead to 6.3\% and 34.0\% increase in total sterilization counts compared to perfect classification.
We note that perfect classification provides target values for LOS prediction methods for various simulation outputs such as sterilization counts and average number of patients in GW/SSU.
\tcolB{Considering the trade offs between different simulation outputs (e.g., average number of patients in GW is inversely proportional to those in SSU), the prediction model that performs similar to perfect classification might be preferrable over others.
In the infinite capacity case, we find that LGBM\_r and LGBM\_p perform closest to perfect classification in terms of total sterilization counts and unit occupation statistics, respectively.}

\setlength{\tabcolsep}{4.5pt}
\renewcommand{\arraystretch}{1.2}
\begin{table}[!h]
    \centering
    \caption{Summary of simulation results with infinite capacity system. Total patient count: 4,597, SS: 1,557, LS: 3,040}
    \begin{footnotesize}
    \scalebox{0.85}{
    \-\hspace{-0.6cm}\input{tables/MLRess_revision2/des_results_GW_CAPACITY_inf_small.tex}
    }
    \end{footnotesize}
    \label{tab:simRessInfCap}
\end{table}

Table~\ref{tab:simRessCapacitated} provides simulation results for the capacitated hospital case, which better reflects the real clinical practice.
Here, we also include GW and SSU ``$\geq 90\%$ utilization rate'' (referred to as ``high utilization rate''), which corresponds to the percentage of time the units are occupied at 90\% or higher capacity.
We first compare the cases where there are no SSU (SSU: 0) and 10 SSUs (SSU: 10) using perfect classification, which makes it possible to directly assess the impact of SSUs without accounting for prediction model accuracy.
We find that having an additional 10 SSUs would reduce sterilization counts by 15.0\%, average GW bed wait times by 24.8\%, average number of patients in WA by 24.3\%, and GW utilization rates by 7.4\%.
In addition, we observe that, for the SSU capacity of 10 units, the perfect classification method leads to 13.2\% SSU high utilization rate, and 6.1 patients on average in the SSUs.
We note that the random classification leads to a lower average waiting times for GW beds, mainly because a large number of LS patients are assigned to SSU beds, causing a significant increase in congestion for SSU. 
In addition, the highest number of sterilization counts are observed for random classification, driven by the increased number of sterilizations in GW and SSU caused by misclassifications.
There are various other factors contributing to the higher GW wait times for higher accuracy classification models (e.g., perfect classification and LGBM). 
First of all, misclassified patients do not contribute to waiting time for GW beds until they are placed in GW queue. 
In addition, having SS patients in GW (i.e., due to misclassification) might contribute to a higher rate of GW beds becoming available over time, and hence causing the average waiting time to be shorter in case of misclassifications. 
Accordingly, perfect classification and LGBM tend to have higher waiting times when compared to random classification. 

\tcolB{Several key simulation outputs for LGBM are found to be between those of LGBM\_r and LGBM\_p.
Compared to the infinite capacity case (see Table~\ref{tab:simRessInfCap}), the highest increase in sterilization counts is observed for LGBM\_r, which is driven by the increase in sterilization counts for WA beds. 
High number of FP (i.e., SS being labeled as LS) causes congestion in GW, increasing the sterilization counts for both WA and GW beds. 
Similarly high average waiting times for GW beds in LGBM\_r can be attributed to high number of SS patients occupying GW beds.
Closest results to those of perfect classification is observed by LGBM\_p, indicating that a high precision model (which lead to low FP --SS being mislabeled as LS--, high FN --LS being mislabeled as SS--) is preferable for LOS prediction given the amount of LS/SS patient arrivals in our dataset as well as GW/SSU capacities.}

\setlength{\tabcolsep}{6pt}
\begin{table}[!ht]
\begin{center}
\caption{Summary of simulation results with the capacitated system. Total patient count: 4,597, SS: 1,557, LS: 3,040}
    \begin{footnotesize}
    \scalebox{0.85}{
    \-\hspace{-0.8cm}\input{tables/MLRess_revision2/des_results_GW_CAPACITY_70_small.tex}
    }
    \end{footnotesize}
    \label{tab:simRessCapacitated}
\end{center}
\end{table}

\subsubsection{Incremental analysis with hospital unit capacities}
\begin{figure}[!ht]
    \centering
    \subfloat[Avg. wait time for GW beds]{\includegraphics[width=0.849\textwidth]{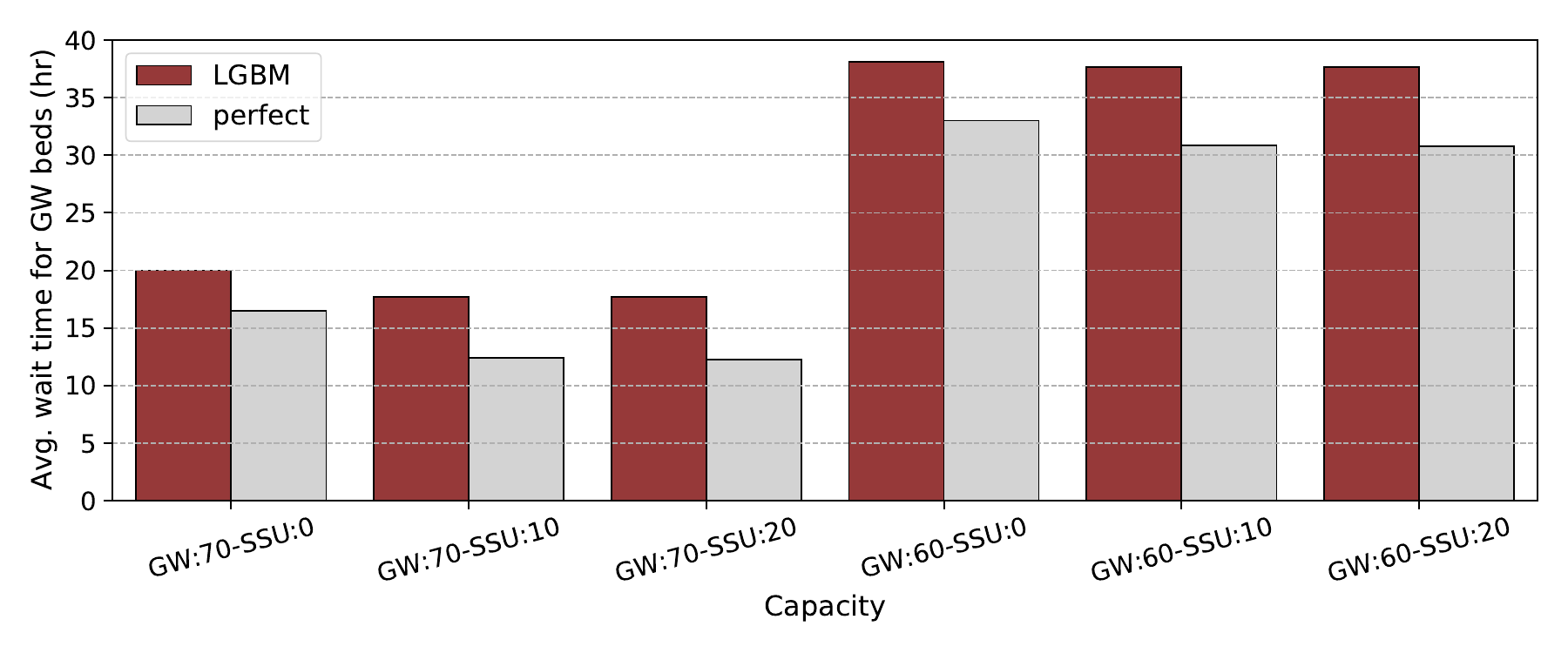}}
    \hfill
    \subfloat[Total sterilization counts]{\includegraphics[width=0.849\textwidth]{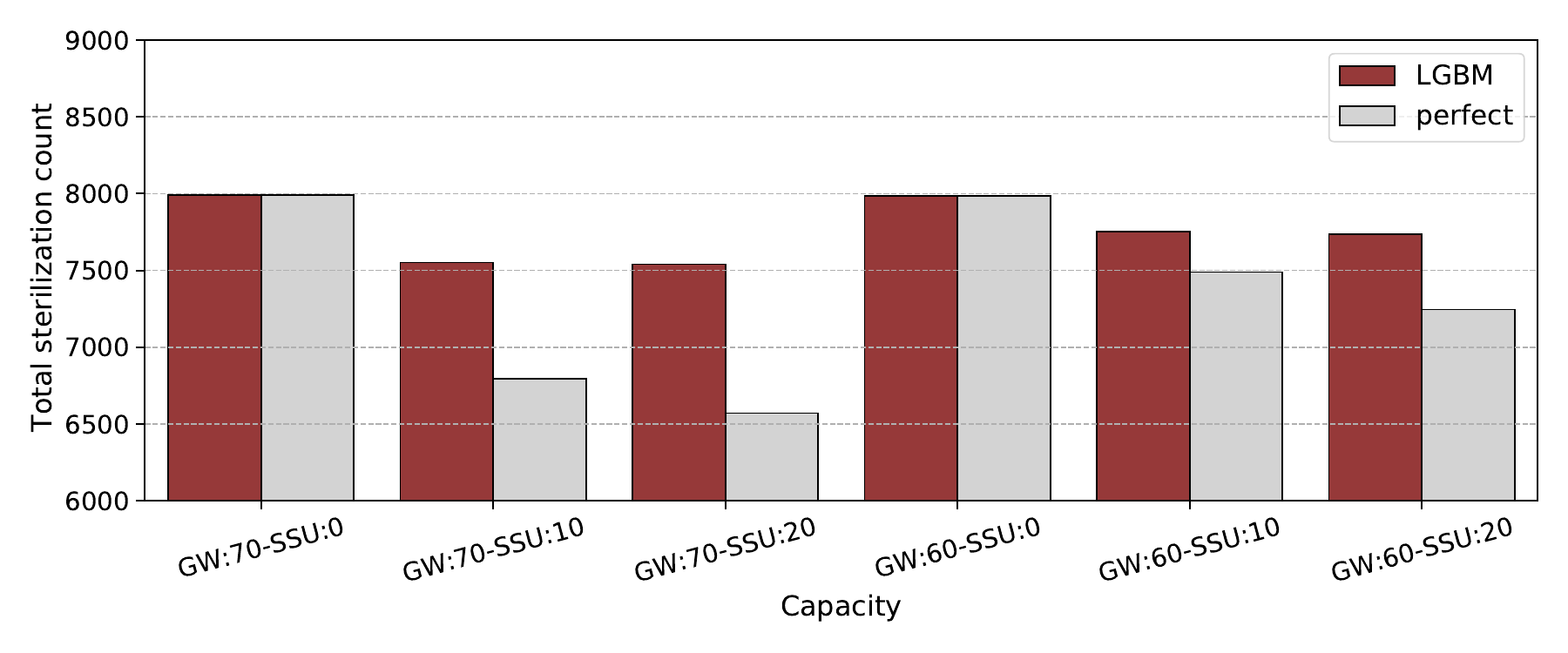}}
    \caption{Performance values for perfect classification and LGBM under different capacity constraints}
    \label{fig:CapPerformanceComparison}
\end{figure}

Figure~\ref{fig:CapPerformanceComparison} shows the results obtained by perfect classification and LGBM under different capacity considerations. 
We observe that introducing SSU improves the system in terms of average GW bed wait times and total sterilization counts. 
The improvement in GW wait times is expected since extra capacity is introduced to the system. 
These results indicate that incremental benefit of increasing SSU capacity (i.e., going from 10 SSU beds to 20 SSU beds) is relatively low. 
That is, both for the LGBM and perfect classification cases, increasing the SSU capacity from 10 to 20 marginally improves the average GW bed wait times.


\tcolB{Our simulation model assumes that during the hospital stay of a patient, if no bed becomes available, then the patient will be discharged from the waiting area.
This assumption is particularly relevant for short stay patients due to their relatively short LOS values.
If a GW bed becomes available before an SS patient is discharged, then the patient is moved to GW, causing two room sterilizations.
Accordingly, for perfect classification, when the capacity increases from (GW=70, SSU=0) to (GW=70, SSU=10), then some of the SS patients who would normally be discharged from WA due to not having SSUs are assigned to SSU beds after a short amount of time that they spend in WA.
Therefore, these SS patients cause a sterilization in SSU, and increase SSU+GW sterilization counts (i.e., +264 sterilizations), which can also be observed in Table~\ref{tab:simRessCapacitated}.
On the other hand, more patients will skip WA due to availability of SSUs. 
Specifically, SS patients can be directly assigned to SSU beds without occupying WA beds, and LS patients would also skip WA at a higher rate since more GW beds become available as SS patients are assigned to SSUs.
As such, WA sterilization counts decrease at a higher rate, reducing overall sterilization counts (i.e., -1,196 sterilizations).}

We can make similar observations about perfect classification case sterilization counts when capacity increases from (GW=70, SSU=10) to (GW=70, SSU=20). 
However, incremental increase in SSU+GW sterilization counts (i.e., +4 sterilizations) and decrease in total sterilization counts (i.e., -223 sterilizations) are smaller compared to the one observed when SSU capacity goes from 0 to 10.
This result implies that, when GW capacity is 70, the SSU capacity of 10 seems to be sufficient for our dataset to accommodate the significant majority of the SS patients.
In addition, GW capacity of 60 also leads to similar observations for sterilization counts regarding the impact of increasing SSU capacity.
We observe that while the total sterilization counts are at the same level when SSU capacity is 0, smaller GW capacity causes a substantial increase in total sterilization counts for identical SSU capacities.
This can be attributed to the fact that more SS/LS patients would cause double sterilizations (i.e., one at WA and another one at GW) as fewer patients would have a chance to skip WA.

\tcolB{While we observe that increasing SSU capacity from 10 to 20 benefits reducing the total sterilization counts in the case of perfect classification, it leads to marginal improvements when LGBM is employed.
That is, since LGBM misclassifies a significant portion of SS patients, it cannot take full advantage of SSUs, and those patients first lead to a sterilization in WA (e.g., if GW is full), which might be followed up by a sterilization in GW.}

\subsubsection{Impact of prediction performance metrics on simulation outcomes}

\begin{figure}[!ht]
    \centering
    \subfloat[Avg. wait time for GW beds]{\includegraphics[width=0.495\textwidth]{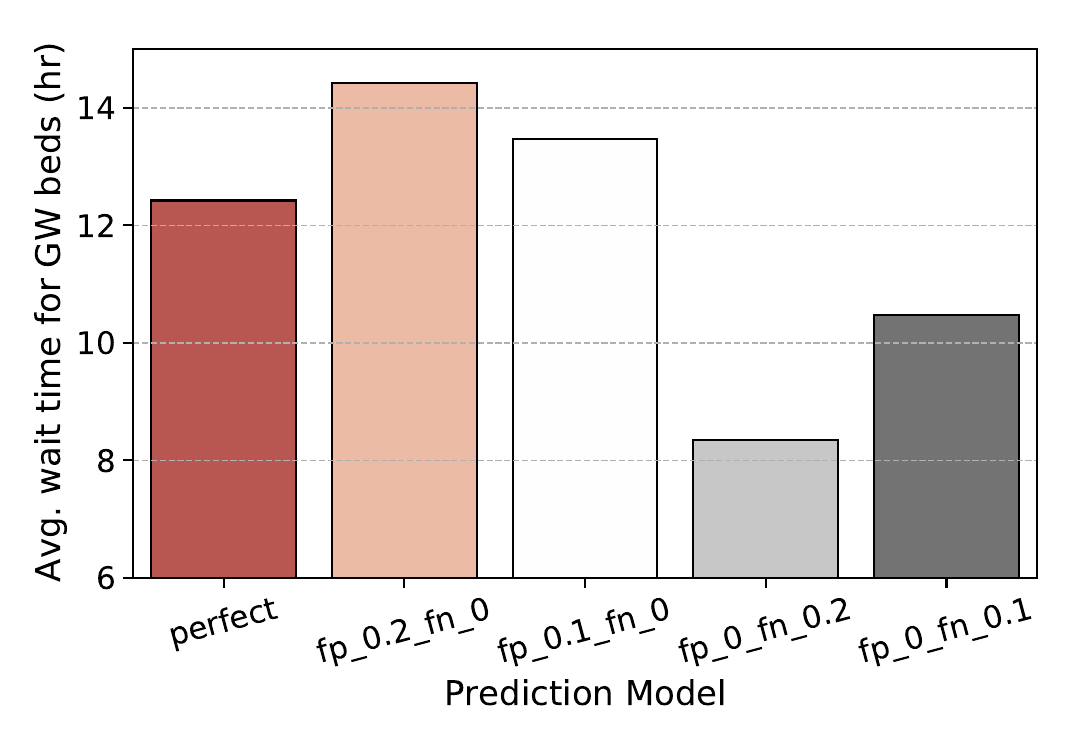}}
    \hfill
    \subfloat[Total sterilization count]{\includegraphics[width=0.495\textwidth]{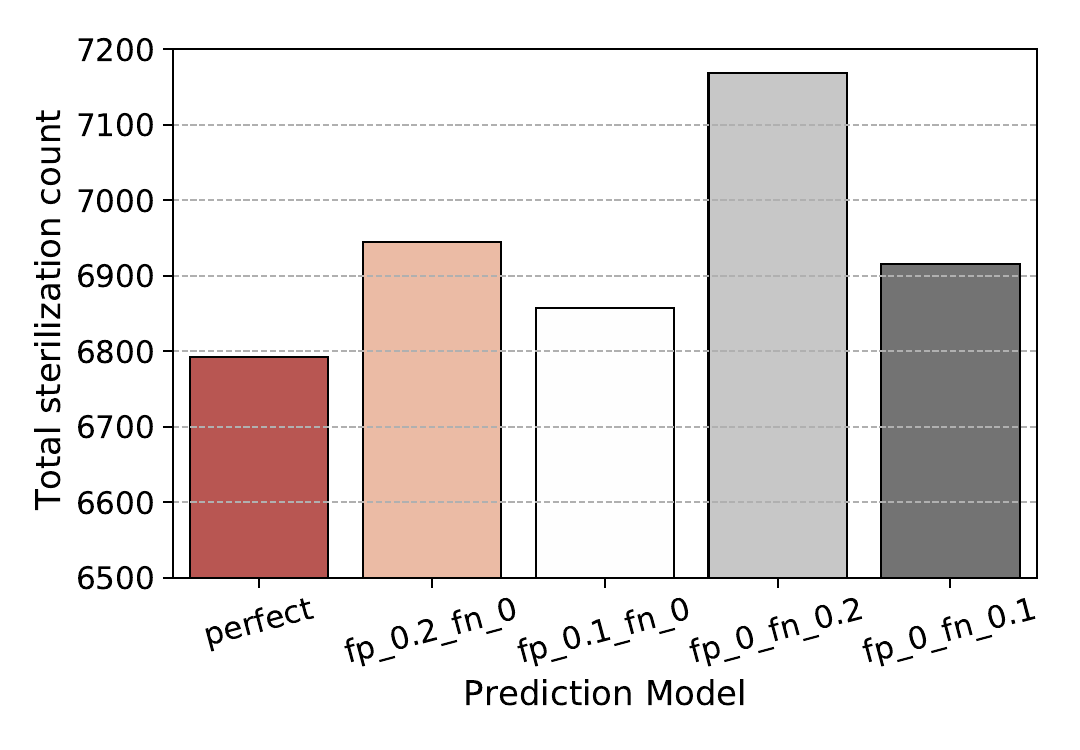}}
    \caption{Performance comparison for different hypothetical prediction models (GW capacity: 70, SSU capacity: 10)}
    \label{fig:SimPerformanceComparisonHypothetical}
\end{figure}

Lastly, we consider hypothetical models with varying levels of performances to better understand which prediction performance metric is more impactful for LOS prediction. 
For simplicity, instead of experimenting with different recall and precision values, we only consider FN and FP rates, which can be taken as surrogates for recall and precision.
Figure~\ref{fig:SimPerformanceComparisonHypothetical} shows that FN rate (as can be seen by the cases fp\_0\_fn\_0.2 --FP rate 0\% and FN rate 20\%-- and fp\_0\_fn\_0.1--FP rate 0\% and FN rate 10\%) 
has more impact on total sterilization count performance metric, and same level of FN rate leads to a higher number of sterilizations compared to its FP counterpart.
This result is expected since a high FN rate implies a high number of LS (SS) patients being labeled as SS (LS), causing more LS (SS) patients to occupy SSU (GW) beds, causing increase in the sterilization counts. 
\tcolB{In addition, while the models fp\_0.1\_fn\_0 and fp\_0.2\_fn\_0 do not lead to any LS patient to be misclassified as SS (i.e., 0 FNs), we still observe increase in total sterilization counts compared to perfect classification since misclassification of LS patients implies many SS patients to occupy the GW beds.
This would decrease the utilization rates for SSUs, and cause more LS patients to have two sterilizations (one at WA and another one at GW) due to fully occupied GW.
For instance, we observe that SSU $\geq 90\%$ utilization rates drop from 13.2\% to 6.3\%, and GW $\geq 90\%$ utilization rates increase from 85.4\% to 86.6\% when we compare perfect classification to fp\_0.2\_fn\_0, indicating the impact of increased occupancy rates on total sterilization counts.}

Regarding the average GW bed wait times, 
we note that increased FP rates leads to higher number of SS patients being assigned to GW instead of SSU, causing increase in wait times. 
On the other hand, increased FN rates lead to LS patients being assigned to SSUs, which yields a decrease in the wait times (because a portion of their time is spent in SSU rather than waiting area).
Accordingly, high FN rates would be preferrable over high FP rates when the primary objective is to reduce average wait times for GW beds.

\section{Discussion and Conclusion}\label{sec:conclusion}
In this study, we developed an extensive generic machine learning framework that allowed us to mix and match different pre-processing, feature engineering, algorithms and visualization techniques using python programming language and scikit-learn package \citep{Pedregosa2011}. Using this framework, we were able to conduct a large number of experiments, which in turn allowed us to gain deeper insights about the large number of features in a systematic fashion and contributed to our search for the best approach.

In our analysis, we used all the available data until admission time to predict the LOS of the patient after admission. That included any previous visits to the ED of the same patient. Specifically, since the encounters in the dataset included a hash code that linked the encounters of a single patient, we were able to gather data from previous visits. Our experiments with various feature selection/engineering approaches provided valuable insights regarding the problem characteristics and helped improve the LOS prediction performance. Our study showed that boosted trees worked best for predicting the patient type (i.e., LS and SS patients) whereas different feature selection strategies did not substantially impact the model performance. On the other hand, we observed that feature engineering was helpful, especially including data from previous admissions. We further note that a prediction model with reasonably good performance could be developed from routinely available data. Even when model predictions were wrong, the models still provided useful information, e.g., predicted SS patients had shorter LOS on average. 

Data originating from a magnitude of different sources can be used to address the LOS prediction problem, including patient history, current patient condition and hospital services quality. Each piece of information contributes to the total number of features available. For instance our comprehensive dataset contains 1,277 number of distinct features, which is a large number of features to work with. 
In such settings, success of the prediction algorithms are greatly effected by the number of available data points. 
On the other hand, our experiments showed that, despite the abundance of features, our dataset lacked highly informative features on LOS values (i.e., class label), which adversely affected prediction performance.

LOS prediction can be a challenging problem for physicians as well. \citet{dent2007can} conduct a study in St. Vincent’s Hospital, Melbourne, investigating physicians' performance in identifying short stay ($\leq 3$ days) patients. They report that only 46\% (122/268; CI 40\%–52\%) of doctors’ predictions are correct. In the same study authors report a 35\% (361/1024) prediction accuracy in algorithmic LOS prediction. Thus, the overall AUC value of 69\% obtained through the LGBM model is likely better than previously reported physician prediction accuracy \citep{dent2007can}. The prediction models provide opportunities for automation and streamlining the hospital operations, however, their adoption largely depends the overall prediction performance. While our study provides evidence on the use of machine learning models for LOS prediction, further research and data collection are needed to come up with models that are suitable for large-scale adoption of such models in hospital settings.

We also designed a simulation model to evaluate the impact of LOS classification in a hospital setting. This evaluation method is not a commonly used approach in applied machine learning research, where the model performance is mostly assessed through an independent test set, and real-life implications of the predictive modelling are not explicitly addressed. Therefore, our adopted approach distinguishes our study from similar studies on LOS prediction in hospital settings. Our simulation model helps quantifying the impact of the high accuracy LOS classification models with respect to several metrics such as bed sterilization counts and average time spent by the patients in GW and SSU. The simulation model could also be tailored to local contexts. For instance, our simulation model showed that improved prediction performance was less impactful in a scenario where patient volume approached hospital capacity (i.e., capacitated scenario), because beds were often occupied, which minimized the ability to efficiently allocate resources to optimize resource usage. 

This study has several limitations. First, the data and results are obtained from a single academic hospital and our specific findings may not be generalizable to other settings. However, we have treated this study as a prototype, which demonstrates a methodology for predicting LOS and simulating the impact of improved predictions on the efficiency of resource usage. Our methods could be applied to other settings. Second, we used only routinely collected data that were available in hospital electronic medical records. These data do not include important information that is likely to influence LOS, such as patient functional status, housing, and caregiver supports. Third, our focus was on predicting LOS, not clinical outcomes or severity of illness. It is possible that patients who were predicted to have long hospital LOS would die in hospital and shorten LOS, leading to misclassification of the prediction model. We did not account for the competing risk of death in this study, because our interest was on predicting hospital resource use not clinical outcomes.



In this study, involving more than 16,000 encounters over 5 years at one hospital, we describe a methodological approach to predicting hospital LOS and a simulation model to evaluate LOS prediction models. 
Our study offers methodological insights for future research in this area, including approaches to feature engineering, feature selection, model selection, and simulation. Finally, our study offers clinical insights, highlighting that the impact of prediction-based bed allocation may be less effective when hospitals are already operating near capacity. Future research directions include validating our findings in other hospitals and improving the simulation model by considering different sequencing decisions regarding patient bed allocation, as well as extending the simulation model by incorporating admissions from other departments in the hospital to GW. In particular, building a prediction model using a dataset obtained from one hospital and testing the model performance on another hospital would provide valuable insights on LOS prediction and its clinical implications.


\section*{Acknowledgements}
The authors would like to thank Saeede S. Asadi K. and Ceni Babaoglu for valuable discussions and contributions throughout this work.

\singlespacing
\bibliographystyle{abbrvnat}
\bibliography{sigproc}

\end{document}

%% file: tables/MLRess_revision2/confusion.tex
\begin{tabular}{lrrrr}
\toprule
& TP    & TN    &  FP   & FN \\
\midrule
LDA - count     & 2,389	& 675	& 882	& 651 \\
LGBM - count     & 2,752	& 491	& 1,066	& 288 \\
LGBM\_r - count     & 2,967	& 175	& 1,382	& 73 \\
LGBM\_p - count     & 2,513	& 678	& 879	& 527 \\
\midrule
LGBM - actual mean LOS (hr)  & 262.8	& 41.7	& 45.5	& 176.0 \\
LGBM - actual median LOS (hr)& 167.0	& 42.0	& 45.0	& 134.5 \\
LGBM - actual LOS $\sigma$ (hr)& 309.6	& 17.4	& 17.6	& 168.1 \\
\midrule
\end{tabular}

%% file: tables/MLRess_revision2/tbr_LDA.tex
\newcommand{\mcp}{\multicolumn{1}{p{1.95cm}}}
\newcommand{\mcps}{\multicolumn{1}{p{1.3cm}}}
\newcommand{\mro}{\multirow{2}{3.6cm}}
\newcommand{\mros}{\multirow{2}{1.3cm}}
\newcommand{\mrt}{\multirow{2}{0.7cm}}
\newcommand{\cnt}{\centering}
\newcommand{\rrt}{\raggedright}
\begin{tabular}{lcccccccc}
\toprule
\mro{\rrt FS method} & \mros{\rrt \# of\\ features} & \mcps{\cnt F1-score} & \mcps{\cnt F1-score\\ std.} & \mcps{\cnt Accuracy\\} & \mcps{\cnt Precision\\ (LS)} & \mcps{\cnt Precision\\ (SS)} & \mcps{\cnt Recall\\ (LS)} & \mcps{\cnt Recall\\ (SS)} \\
\midrule
No FS                & 114 & 0.659 & 0.011 & 0.663 & 0.735 & 0.499 & 0.769 & 0.453 \\
feature\_importances & 40  & 0.637 & 0.015 & 0.627 & 0.768 & 0.460 & 0.629 & 0.624 \\
Ensemble         & 40  & 0.640 & 0.014 & 0.633 & 0.750 & 0.462 & 0.670 & 0.560 \\
feature\_importances & 15  & 0.630 & 0.015 & 0.620 & 0.761 & 0.452 & 0.624 & 0.613 \\
Ensemble         & 15  & 0.625 & 0.014 & 0.616 & 0.746 & 0.444 & 0.638 & 0.572 \\
feature\_importances & 5   & 0.613 & 0.012 & 0.602 & 0.748 & 0.434 & 0.604 & 0.599 \\
Ensemble         & 5   & 0.623 & 0.012 & 0.613 & 0.752 & 0.444 & 0.623 & 0.594 \\
\bottomrule
\end{tabular}

%% file: tables/MLRess_revision2/tbr_LGBM.tex
\newcommand{\mcp}{\multicolumn{1}{p{1.95cm}}}
\newcommand{\mcps}{\multicolumn{1}{p{1.3cm}}}
\newcommand{\mro}{\multirow{2}{3.6cm}}
\newcommand{\mros}{\multirow{2}{1.3cm}}
\newcommand{\mrt}{\multirow{2}{0.7cm}}
\newcommand{\cnt}{\centering}
\newcommand{\rrt}{\raggedright}
\begin{tabular}{lcccccccc}
\toprule
\mro{\rrt FS method} & \mros{\rrt \# of\\ features} & \mcps{\cnt F1-score} & \mcps{\cnt F1-score\\ std.} & \mcps{\cnt Accuracy\\} & \mcps{\cnt Precision\\ (LS)} & \mcps{\cnt Precision\\ (SS)} & \mcps{\cnt Recall\\ (LS)} & \mcps{\cnt Recall\\ (SS)} \\
\midrule
No FS                & 114 & 0.664 & 0.009 & 0.658 & 0.766 & 0.493 & 0.698 & 0.579 \\
feature\_importances & 40  & 0.658 & 0.011 & 0.649 & 0.777 & 0.483 & 0.663 & 0.624 \\
Ensemble         & 40  & 0.646 & 0.009 & 0.637 & 0.769 & 0.470 & 0.647 & 0.617 \\
feature\_importances & 15  & 0.644 & 0.012 & 0.634 & 0.774 & 0.468 & 0.635 & 0.633 \\
Ensemble         & 15  & 0.590 & 0.017 & 0.580 & 0.764 & 0.422 & 0.530 & 0.677 \\
feature\_importances & 5   & 0.613 & 0.011 & 0.602 & 0.763 & 0.438 & 0.582 & 0.643 \\
Ensemble         & 5   & 0.573 & 0.014 & 0.564 & 0.765 & 0.412 & 0.495 & 0.700 \\
\bottomrule
\end{tabular}

%% file: tables/MLRess_revision2/des_results_GW_CAPACITY_inf_small.tex
\begin{tabular}{lrrrrr}
\toprule
\textbf{Prediction model}                      & \textbf{perfect} & \textbf{random} & \textbf{LGBM}  & \textbf{LGBM\_r} & \textbf{LGBM\_p} \\
\midrule
total sterilization count             & 4,595   & 6,158  & 4,883 & 4,668   & 5,122   \\
GW sterilization count        & 3,039   & 3,806  & 4,105 & 4,421   & 3,918   \\
SSU sterilization count       & 1,556   & 2,352  & 778   & 247     & 1,204   \\
WA sterilization count     & -       & -      & -     & -       & -       \\
\midrule
avg. wait time for GW bed (hr)       & -       & -      & -     & -       & -       \\
\midrule
avg. LS patient time in GW (hr) & 254.6   & 229.3  & 249.9 & 253.4   & 246.1   \\
avg. LS patient time in SSU (hr) & -       & 72     & 72    & 72      & 72      \\
avg. SS patient time in GW (hr) & -     & 44.5   & 45.5  & 44.7    & 46.0    \\
avg. SS patient time in SSU (hr) & 44.3    & 44.0   & 41.6  & 41.0    & 42.1    \\
\midrule
max. \# of patients in GW                   & 95      & 91     & 101   & 104     & 96      \\
max. \# of patients in SSU                  & 21      & 22     & 13    & 8       & 17      \\
avg. \# of patients in GW                   & 71.5    & 67.7   & 74.8  & 77.0    & 72.9    \\
avg. \# of patients in SSU                  & 6.6     & 10.6   & 3.3   & 1.0     & 5.2    \\
\midrule
LS misclassification count        & -       & 1,563  & 288   & 73      & 527     \\
SS misclassification count       & -       & 767    & 1,066 & 1,382   & 879     \\
\bottomrule
\end{tabular}

%% file: tables/MLRess_revision2/des_results_GW_CAPACITY_70_small.tex
\newcommand{\mcp}{\multicolumn{1}{p{1.95cm}}}
\newcommand{\mcps}{\multicolumn{1}{p{1.3cm}}}
\newcommand{\mro}{\multirow{2}{3.6cm}}
\newcommand{\mros}{\multirow{2}{1.3cm}}
\newcommand{\mrt}{\multirow{2}{0.7cm}}
\newcommand{\cnt}{\centering}
\newcommand{\rrt}{\raggedright}
\newcommand{\rlt}{\raggedleft}
\begin{tabular}{lrrrrrr}
\toprule
\textbf{Prediction model} & \mcps{\rlt \textbf{perfect}\\ \textbf{SSU:0}} & \mcps{\rlt \textbf{perfect}\\ \textbf{SSU:10}} & \mcps{\rlt \textbf{random}\\ \textbf{SSU:10}} & \mcps{\rlt \textbf{LGBM}\\ \textbf{SSU:10}}  & \mcps{\rlt \textbf{LGBM\_r}\\ \textbf{SSU:10}} & \mcps{\rlt \textbf{LGBM\_p}\\ \textbf{SSU:10}} \\
\midrule
total sterilization count & 7,989 & 6,793 & 7,900 & 7,551 & 7,846 & 7,395 \\
GW sterilization count & 4,226 & 3,109 & 3,854 & 3,912 & 4,112 & 3,792 \\
SSU sterilization count & - & 1,481 & 1,869 & 773 & 247 & 1,163 \\
WA sterilization count & 3,763 & 2,203 & 2,177 & 2,866 & 3,487 & 2,440 \\
\midrule
avg. wait time for GW bed (hr) & 16.5 & 12.4 & 7.5 & 17.7 & 21.9 & 14.6 \\
\midrule
avg. LS patient time in GW (hr) & 229.7 & 235.8 & 224.6 & 229.9 & 229.9 & 229.1 \\
avg. LS patient time in SSU (hr) & - & - & 53.0 & 52.9 & 52.8 & 52.9 \\
avg. SS patient time in GW (hr) & 28.8 & 39.4 & 35.3 & 31.8 & 30.4 & 33.9 \\
avg. SS patient time in SSU (hr) & - & 43.2 & 40.5 & 41.6 & 41.0 & 41.8 \\
\midrule
avg. \# of patients in GW & 67.8 & 66.5 & 65.8 & 67.3 & 67.7 & 66.8 \\
avg. \# of patients in SSU & - & 6.1 & 8.7 & 3.4 & 1.0 & 5.2 \\
avg. \# of patients in WA & 7.4 & 5.6 & 3.0 & 7.9 & 9.8 & 6.4 \\
\midrule
GW $\geq$ 90\% utilization rate & 91.7\% & 85.4\% & 80.9\% & 88.5\% & 91.1\% & 86.3\% \\
SSU $\geq$ 90\% utilization rate & - & 13.2\% & 51.1\% & 1.6\% & 0.0\% & 10.0\% \\
\bottomrule
\end{tabular}
